%% file: asyncBO_paper.tex
\documentclass{article}

\usepackage{microtype}
\usepackage{graphicx}
\usepackage{subcaption}

\usepackage{booktabs} 
\usepackage{amsmath}
\usepackage{amsfonts}
\usepackage{amssymb}
\DeclareMathOperator*{\argmax}{arg\,max}

\usepackage[dvipsnames]{xcolor}
\usepackage{hyperref}
\usepackage{multirow}

\usepackage[accepted]{icml2019}
\icmltitlerunning{Asynchronous Batch Bayesian Optimisation with Improved Local Penalisation}

\begin{document}

\twocolumn[
\icmltitle{Asynchronous Batch Bayesian Optimisation with Improved Local Penalisation}

\icmlsetsymbol{equal}{*}

\begin{icmlauthorlist}
\icmlauthor{Ahsan S. Alvi}{equal,to,mf}
\icmlauthor{Binxin Ru}{equal,to}
\icmlauthor{Jan Calliess}{to,om}
\icmlauthor{Stephen J. Roberts}{to,mf,om}
\icmlauthor{Michael A. Osborne}{to,mf}
\end{icmlauthorlist}

\icmlaffiliation{to}{Department of Engineering Science, University of Oxford}
\icmlaffiliation{mf}{Mind Foundry Ltd., Oxford, UK}
\icmlaffiliation{om}{Oxford-Man Institute of Quantitative Finance}

\icmlcorrespondingauthor{Binxin Ru}{robin@robots.ox.ac.uk}
\icmlcorrespondingauthor{Ahsan Alvi}{asa@robots.ox.ac.uk}

\icmlkeywords{Machine Learning, ICML}

\vskip 0.3in
]

\printAffiliationsAndNotice{\icmlEqualContribution} 
\begin{abstract}
Batch Bayesian optimisation (BO) has been successfully applied to hyperparameter tuning using parallel computing, but it is wasteful of resources: workers that complete jobs ahead of others are left idle. We address this problem by developing an approach, \emph{Penalising Locally for Asynchronous Bayesian Optimisation on $k$ workers} (PLAyBOOK), for asynchronous parallel BO. We demonstrate empirically the efficacy of PLAyBOOK and its variants on synthetic tasks and a real-world problem. We undertake a comparison between synchronous and asynchronous BO, and show that asynchronous BO often outperforms synchronous batch BO in both wall-clock time and number of function evaluations.
\end{abstract}


\section{Introduction}
Bayesian optimisation (BO) is a popular sequential global optimisation technique for functions that are expensive to evaluate \cite{brochu2010tutorial}. 
Whilst standard BO may be sufficient for many applications, it is often the case that multiple experiments can be run at the same time in parallel.
For example, in the case of drug discovery, many different compounds can be tested in parallel via high throughput screening equipment \cite{hernandez2017parallel}, and when optimising machine learning algorithms, we can train different model configurations concurrently on multiple workers \cite{chen2018bayesian, kandasamy2018parallelised}. 
This observation lead to the development of parallel (``batch'') BO algorithms, which, at each optimisation step, recommend a batch of $k$ configurations to be evaluated.

In cases where the runtimes of tasks are roughly equal, this is usually sufficient, but, if the runtimes in a batch vary, this will lead to inefficient utilisation of our hardware.
For example, consider the optimisation of the number of units in the layers of a neural network. Training and evaluating a large network (greater number of units per layer) will take significantly longer than a small network (fewer units per layer), so for an iteration of (synchronous) batch BO to complete, we need to wait for the slowest configuration in the batch to finish, leaving the other workers idle. 
In order to improve the utilisation of parallel computing resources, we can run function evaluations \emph{asynchronously}: as soon as $c$ workers ($c < k$) complete their jobs, we choose new tasks for them. 

Although asynchronous batch BO has a clear advantage over synchronous batch BO in terms of wall-clock time \cite{kandasamy2018parallelised}, it may lose out in terms of sample efficiency, as an asynchronous method takes decisions with less data than its synchronous counterpart at each stage of the optimisation.
We investigate this empirically in this work.

Our contributions can be summarised as follows.
\begin{itemize}    
    \item We develop a new approach to asynchronous parallel BO, \emph{Penalising Locally for Asynchronous Bayesian Optimisation on $k$ workers} (PLAyBOOK), which uses penalisation-based strategies to prevent redundant batch selection. We show that our approach compares favourably against existing asynchronous methods. 
       
    \item We propose a new penalisation function, which prevents redundant samples from being chosen. We also propose designing the penalisers using local (instead of global) variability features of the surrogate to more effectively explore the search space.
    
    \item We demonstrate empirically that asynchronous methods perform at least as well as their synchronous variants. We also show that PLAyBOOK outperforms its synchronous variants \emph{both} in terms of wall-clock time and sample efficiency, particularly for larger batch sizes. This renders PLAyBOOK a competitive parallel BO method.
\end{itemize}


\section{Related work}
\label{sec:prior_work}

Many different synchronous batch BO methods have been proposed over the past years. Approaches include using \\hallucinated observations \citep{ginsbourger2010kriging, desautels2014parallelizing}, knowledge gradient~\cite{wu2016parallel}, Determinental point processes \cite{kathuria2016batched}, 
maximising the information gained about the objective function or the global minimiser~\cite{contal2013parallel, shah2015parallel}, and sampling-based simulation~\cite{azimi2010batch, kandasamy2018parallelised, hernandez2017parallel}. 
A recent synchronous batch BO method that demonstrated promising empirical results is Local Penalisation (LP)~\cite{gonzalez2016batch}. 
After adding a configuration $x_j$ to the batch, LP penalises the value of the acquisition function in the neighbourhood of $x_j$, encouraging diversity in the batch selection. 

Asynchronous BO has received surprisingly little attention compared to synchronous BO to date. 
\citet{ginsbourger2011dealing} proposed a sampling-based approach that approximately marginalises out the unknown function values at busy locations by taking samples from the posterior at those locations. Due to its reliance on sampling, it suffers from poor scaling, both in batch size and BO steps.

\citet{wang2016parallel} developed an efficient global optimiser (MOE) which estimates the gradient of q-EI, a batch BO method proposed by \citet{ginsbourger2008multi}, and uses it in a stochastic gradient ascent algorithm to solve the prohibitively-expensive maximisation of the q-EI acquisition function, which selects all points in the batch simultaneously. 

A more recent method utilizes Thompson Sampling \cite{kandasamy2018parallelised} (TS) to select new batch points. 
This has the benefit of attractive scaling, since the method minimises samples from the surrogate model's posterior. In the case of a Gaussian process (GP) model, a batch point is placed at the minimum location of a draw from a multivariate Gaussian distribution. 
The disadvantage of TS is that it relies on the uncertainty in the surrogate model to ensure that the batch points are well-distributed in the search space. 
\section{Preliminaries}
\label{sec:prelim}
To perform Bayesian optimisation to find the global minimum of an expensive objective function \(f\), we must first decide on a surrogate model for \(f\). 
Using a Gaussian process (GP) as the surrogate mode is a popular choice, due to the GP's potent function approximation properties and ability to quantify uncertainty. 
A GP is a prior over functions that allows us to encode our prior beliefs about the properties of the function \(f\), such as smoothness and periodicity. 
A comprehensive introduction to GPs can be found in \cite{rasmussen2006gaussian}. 

For a scalar-valued function $f$ defined over a compact space $\mathcal{X}$: $\mathbb{R}^d \to \mathbb{R}$, we define a GP prior over $f$ to be \(\mathcal{GP}(m(x), k(x, x'; \theta))\) where $m(x)$ is the mean function, $k(\cdot,\cdot)$ is a covariance function (also known as the kernel) and \(\theta\) are the hyperparameters of the kernel.
The posterior distribution of the GP at an input \(\tilde{x}\) is Gaussian:
\begin{equation}
    p(f(\tilde{x}) \mid \tilde{x}, \mathcal{D}_s) = \mathcal{N}\big( f(\tilde{x}) ; \mu(\tilde{x}), \sigma^2(\tilde{x}) \big),
\end{equation}
with mean and variance
\begin{align}
    \mu(\tilde{x}) &= k(\tilde{x}, X) K(X,X)^{-1}Y, \\
    \sigma^2(\tilde{x}) &= k(\tilde{x}, \tilde{x}') - k(\tilde{x}, X) K(X,X)^{-1}k(X, \tilde{x}'),
\end{align}
where \(X\) is a matrix with an input location in each row \(\{x_1, x_2, ... , x_N\}\) and \(Y\) is a column vector of the corresponding observations \(\{y_1, y_2, ..., y_N\}\), where $y_i = f(x_i)$. 
The hyperparameters of the model have been dropped in these equations for clarity.

The second choice we make is that of the acquisition function \(\alpha : \mathbb{R}^d \to \mathbb{R}\).
Many different functional forms for \(\alpha(x)\) have been proposed to date \cite{kushner1964new, jones1998efficient, srinivas2009gaussian, hennig2012entropy, hernandez2014predictive, ru2017fast}, each with their relative merits and disadvantages. 
Although our method is applicable to most acquisition functions, we use the popular GP Upper Confidence Bound (UCB) in our experiments \cite{srinivas2009gaussian}. UCB is defined as
\begin{equation}
    \alpha_{\text{UCB}}(x) = \mu(x) + \kappa \sigma(x),
    \label{eq:ucb}
\end{equation}
where \(\mu(x)\) and \(\sigma(x)\) are the mean and standard deviation of the GP posterior and \(\kappa\) is a parameter that controls the trade-off between exploration (visiting unexplored areas in \(\mathcal{X}\)) and exploitation (refining our belief by querying close to previous samples). 
This parameter can be set according to an annealing schedule \cite{srinivas2009gaussian} or fixed to a constant value. 

\section{Asynchronous vs synchronous BO}
\label{sec:async_vs_synch}
\begin{figure}[htb!]
    \centering
    \includegraphics[trim={5cm 6cm 15cm 2.5cm},clip, width=1.0\linewidth]{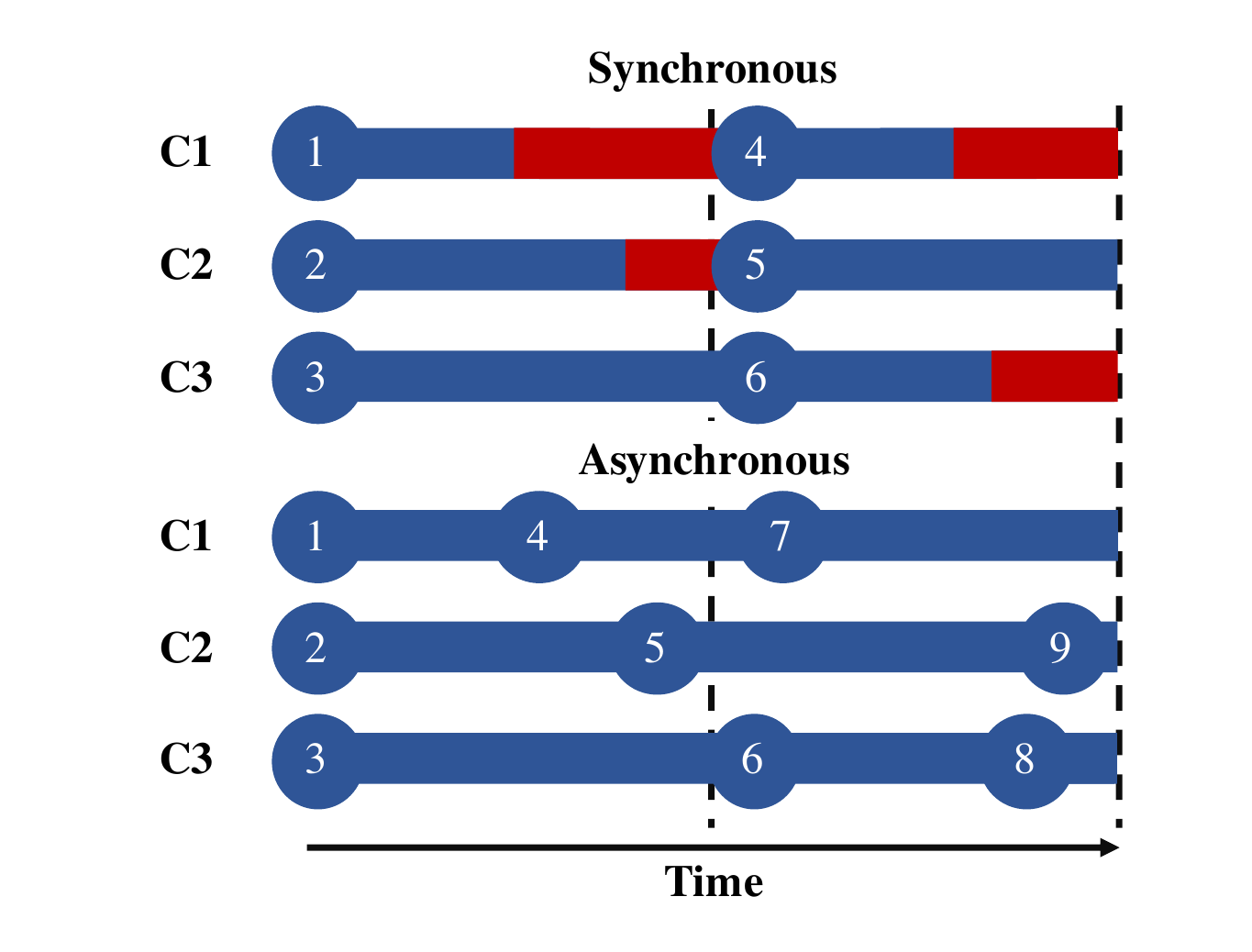}
    \caption{Illustration showing the difference between synchronous and asynchronous batch BO in the case of $k=3$ parallel workers. The blue bar indicates the processing time taken for a worker to evaluate its assigned task and the red bar indicates the waiting time for a worker between completing its previous task and beginning a new task. It is clear that asynchronous batch BO, which makes better use of the computing resources, can complete a greater number of evaluations than its synchronous counterpart within the same duration.}
    \label{fig:sync_vs_async}
\end{figure}

In synchronous BO, the aim is to select a batch of promising locations $\mathcal{B} = \{x_j\}_{j=1}^k$ that will be evaluated in parallel (Fig. \ref{fig:sync_vs_async}). 
Solving this task directly is difficult, which is why most batch BO algorithms convert this selection into a sequential procedure, selecting one point at a time for the batch.
At the $s$th BO step, the optimal choice of batch point $x_{j}$ ($j \in \{1, 2, ..., k\}$) should then not only take into account our current knowledge of \(f\), but also marginalise over possible function values at the locations $\{ x_i \}_{i=1}^{j-1}$  that we have chosen so far for the batch:
\begin{align}
 \label{eq:opt_sync_batch}
    x_j & = \argmax_{x \in \mathcal{X}}
      \int \alpha(x \mid \mathcal{D}_{s}, \mathcal{D}_{j-1}) \nonumber  \\
      & \quad \prod_{i=1}^{j-1}   
     p(y_i \mid x_i, \mathcal{D}_{s}, \mathcal{D}_{i-1}) d y_i,
\end{align}
where $\mathcal{D}_{s}$ are the observations we have gathered so far and $ \mathcal{D}_{j-1} = \{x_i, y_i\}_{i=1}^{j-1}$ and $\mathcal{D}_0 = \varnothing$ \cite{gonzalez2016batch}.

In asynchronous BO, the key motivation is to maximise the utilisation of our $k$ parallel workers. After a desired number of workers $c < k$ complete their tasks, we assign new tasks to them without waiting for the remaining $b = (k-c)$ busy workers to complete their tasks.
Now the general design for selecting the next query point marginalises over the likely function values both at locations under evaluation by busy workers, as well as the already-selected points in the batch:
\begin{align}
 \label{eq:opt_async_batch}
     x_j  =  &\argmax_{x \in \mathcal{X}}
      \iint \alpha(x \mid \mathcal{D}_{s_{-}}, \mathcal{D}_{b}, \mathcal{D}_{j-1} ) \nonumber \\
      & \quad\prod_{i=1}^{j-1}   
     p(y_i \mid x_i, \mathcal{D}_{s_{-}}, \mathcal{D}_{b}, \mathcal{D}_{i-1}) \nonumber \\
     & \quad\prod_{l=1}^{b}   
    p(y_l \mid x_l, \mathcal{D}_{s_{-}}) d y_i d y_l,
\end{align}
where $j \in \{1, ..., c\}$ and $\mathcal{D}_{b}= \{ x_i, y_i \}_{i=1}^{b}$ represents the locations and function values of the busy locations.
$\mathcal{D}_{s_{-}}$ are the observations available at the point of constructing the asynchronous batch.
In general, $\mathcal{D}_{s_{-}}$ contains fewer observations than the $\mathcal{D}_{s}$ that would be used to select the equivalent batch of evaluations in the synchronous setting.  
Fig. \ref{fig:sync_vs_async} shows the case of $c=1$ and thus $b=k-1=2$. 

In a given period of time, asynchronous batch BO is able to process a greater number of evaluations than the synchronous approach: asynchronous BO offers clear gains in resource utilisation. 
However, \citet{kandasamy2018parallelised} claim that the asynchronous setting may not lead to better performance when measured by the number of evaluations. 
The authors point out that a new evaluation in a sequentially-selected synchronous batch will be selected with at most $k-1$ evaluations ``missing'' (that is, with knowledge of their locations $x$ but absent the knowledge of their values $y$), corresponding to the previously-selected points in the current batch (i.e. $j-1 \leq k-1$ in Eq. \eqref{eq:opt_sync_batch}). 
Evaluations in the asynchronous case
are always chosen with $k-1$ ``missing'' evaluations. 

However, to our knowledge, there exists little empirical investigation of the performance difference between synchronous and asynchronous batch methods. 
We conducted this comparison on a large set of benchmark test functions and found that asynchronous batch BO can be as good as synchronous batch BO for different batch selection methods. Additionally, for the penalisation-based methods we propose, asynchronous operation often outperforms the synchronous setting, particularly as the batch size increases. 
We will discuss this interesting empirical observation in Section \ref{subsec:exps-sync-vs-async}.


\section{Penalisation-based asynchronous BO}
\label{sec: async_with_penalisation}

We now present our core algorithmic contributions. 
As discussed in Section \ref{sec:prior_work}, the existing asynchronous BO methods suffer drawbacks such as the prohibitively high cost of repeatedly updating GP surrogates when selecting batch points \cite{ginsbourger2011dealing} or the risk of redundant sampling at or near a busy location in the batch \cite{kandasamy2018parallelised}. In view of these limitations, we propose a penalisation-based asynchronous method which encourages sampling diversity among the points in the batch as well as eliminating the risk of repeated samples in the same batch. Our proposed method remains computationally efficient, and thus scales well to large batch sizes. 

Inspired by the Local Penalisation approach (LP) in synchronous BO \cite{gonzalez2016batch}, we approximate Eq. \eqref{eq:opt_async_batch} for the case of $c=1$ as:
 \begin{equation}
 \label{eq:async_LP}
    x_j = \arg\max_{x \in \mathcal{X}} \left \{ \alpha(x \mid \mathcal{D}_{s_{-}})
    \prod_{i=1}^{k-1}\phi(x \mid x_i , \mathcal{D}_{s_{-}})\right \},
\end{equation}

where $\phi(x \mid x_i , \mathcal{D}_{s_{-}})$ is the penaliser function centred at the busy locations $\{ x_i \}_{i=1}^{k-1}$.
In the following subsections, we design effective penaliser functions by harnessing the Lipschitz properties of the function and its GP posterior. 
To simplify notation, we denote $\phi(x \mid x_i , \mathcal{D}_{s_{-}})$ as $\phi(x \mid x_i)$ and $\alpha(x \mid \mathcal{D}_{s_{-}})$ as $\alpha(x)$ in the remainder of the section.


\subsection{Hard Local Penaliser}
\label{subsubsec:hardPen}
Assume the unknown objective function is Lipschitz continuous with constant $L$ and has a global minimum value $f(x^*) = M$ and $x_j$ is a busy task,
\begin{equation}
  \vert f(x_j) - M \vert \leq L \| x_j - x^* \|.
\end{equation}
This implies $x^*$ cannot lie within the spherical region centred on $x_j$ with radius $r_j =\frac{f(x_j)-M}{L}$: 
\begin{equation}
    \mathbb{S} (x_j,r_j)=\mathcal{X} \ \{ x \in \mathcal{X}: \| x - x_j \| \leq r_j \}.
\end{equation}
If $x_j$ is still under evaluation by a worker, there is no need for any further selections inside $\mathbb{S} (x_j,r_j)$.

Given that $f(x_j) \sim \mathcal{N}(\mu(x_j),\sigma^2(x_j))$ and thus $\mathbb{E}(r_j) = \frac{\vert \mu(x_j)-M \vert}{L}$, applying Hoeffding's inequality for all $\epsilon > 0$ \cite{jalali2013lipschitz} gives
\begin{equation}\label{eq:hoeffd}
    P(r_j > \mathbb{E}(r_j) + \epsilon ) \leq \exp\left(-\frac{2\epsilon^2L^2}{\sigma(x_j)^2}\right),
\end{equation}
which implies there is a high probability (around $99\%$) that $r_j \leq \frac{\vert \mu(x_j)-M \vert}{L} + 1.5 \frac{\sigma(x_j)}{L}$. 

\begin{figure}[t!]
    \centering
    \includegraphics[trim={3.5cm 9.5cm 2.8cm 9cm},clip, width=1.0\linewidth]{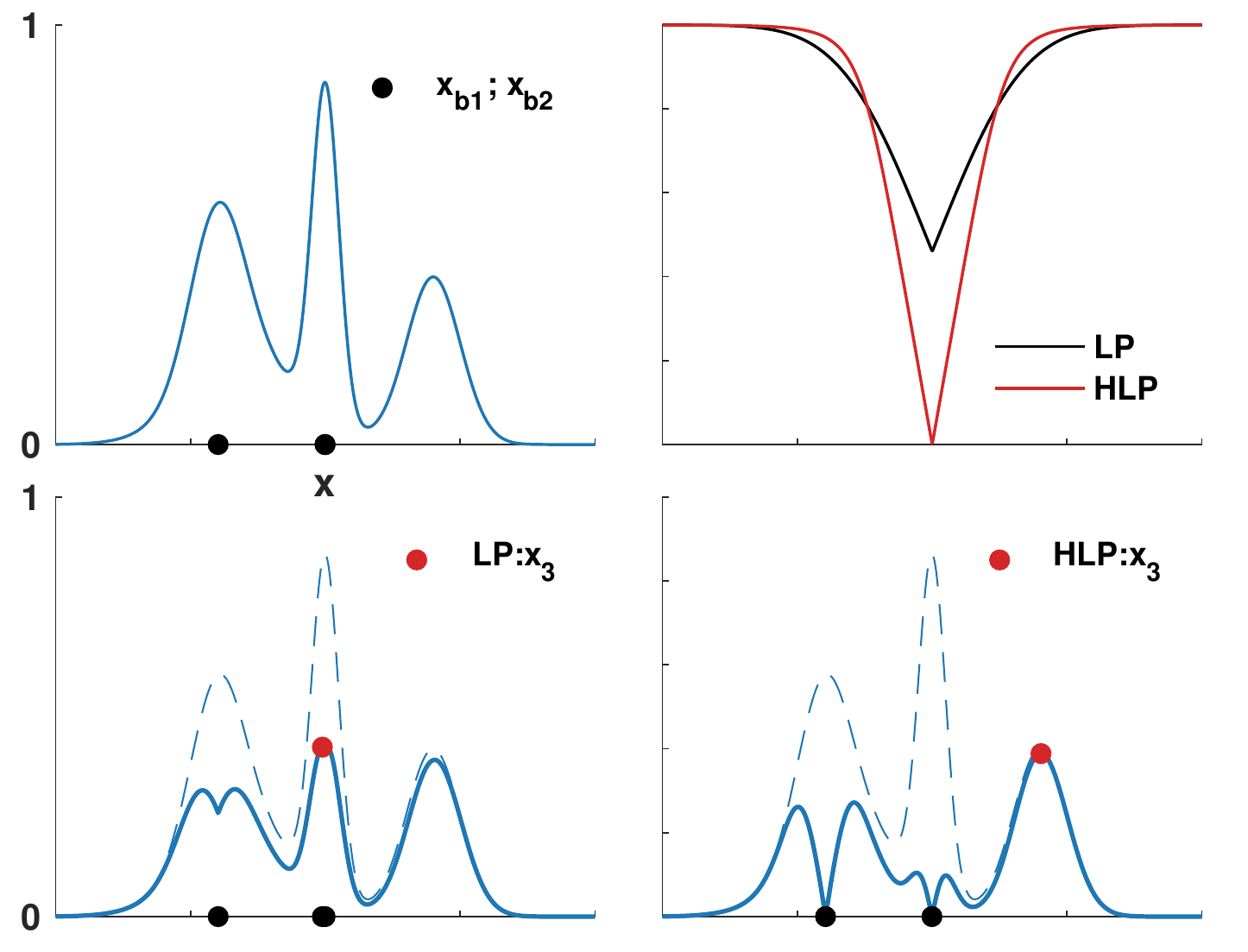}
    \caption{Illustration of asynchronous batch selection by na\"{i}ve LP and HLP. The top left plot shows the acquisition function $\alpha(x)$ and the locations (i.e. $x_{b1}$ and $x_{b2}$ denoted in black dots) under evaluation by busy workers. 
    The top right plot shows the shapes of two penalisers at the busy location $x_{b1}$. Their respective penalisation effects on $\alpha(x)$ at $x_{b1}$ and $x_{b2}$ as well as the new batch point $x_3$ to be assigned to the available worker are shown in the subplots that follow, LP on the left and HLP on the right.}
    \label{fig:lp_vs_hlp}
\end{figure}

The penalisation function $\phi(x \mid x_j)$ should incorporate this belief to guide the selection of the next asynchronous batch point by reducing the value of the acquisition function at locations $\{x \in \mathbb{S} (x_j,r_j) \}$. 
A valid penaliser should possess the several properties:
\begin{itemize}
    \item the penalisation region shrinks as the expected function value at $x_j$ gets close to the global minimum (i.e. small $\vert \mu(x_j) - M \vert$) \cite{gonzalez2016batch};
    \item the penalisation region shrinks as $L$ increases \cite{gonzalez2016batch};
    \item the extent of penalisation on $\alpha(x)$ increases as $x$ gets closer to $x_j$ with $\alpha(x_j)=0$ if $\alpha(x)\geq 0$ for all $x \in \mathcal{X}$.
\end{itemize}
The Local Penaliser (LP) in \cite{gonzalez2016batch} fulfils the first two properties but not the final one which we believe is crucial. Thus, 
directly using it for the asynchronous case makes the algorithm vulnerable to redundant sampling as illustrated in Fig. \ref{fig:lp_vs_hlp}. In view of this limitation, we propose a simple yet effective Hard Local Penaliser (HLP) which satisfies all three conditions
\begin{equation} \label{eq:hlp_init}
    \phi(x \mid x_j) = \min \left\{\frac{\| x- x_j\|}{\mathbb{E}(r_j) + \gamma \frac{\sigma(x_j)}{L}} , 1 \right \},
\end{equation}
where $\gamma$ is a constant.

The above expression can be made differentiable by the approximation:
\begin{equation} \label{eq:hlp}
    \hat{\phi}(x \mid x_j) = \left[ \left(\frac{\| x- x_j\|}{\mathbb{E}(r_j) + \gamma\frac{\sigma(x_j)}{L}}\right)^p + 1^p \right ]^{1/p},
\end{equation}
with $\hat{\phi}(x \mid x_j) \rightarrow \phi(x \mid x_j) $ as $p \rightarrow{-\infty}$.

In addition, the global optimum $M$ is unknown in practice and is usually approximated by the best function value observed $\hat{M}=\min\{f(x_i)\}_i^n$ \cite{gonzalez2016batch}. This approximation tends to lead to underestimation of $\mu(x_j) - M$ and thus $\mathbb{E}(r_j)$, reducing the extent of the penalisation at $x_j$ and in the region nearby. 
HLP mitigates this effect by penalising significantly harder than the penaliser proposed by \citet{gonzalez2016batch}, and maximally at $x_j$ ($\alpha(x_j)=0$). Thus, our method is less affected by over-estimation of the global minimum $\hat{M} > M$.

\subsection{Local Estimated Lipschitz constants}
\label{subsubsec:localL}
In BO, the global Lipschitz constant $L$ of the objective function is unknown. Assuming the true objective function $f$ is a draw from its GP surrogate model, we can approximate $L$ with $\hat{L}= \max_{x \in \mathcal{X}} \| \mu_{\bigtriangledown} (x) \|$ where $ \mu_{\bigtriangledown} (x)$ is the posterior mean of the derivative GP \cite{gonzalez2016batch}. However, using the estimated global Lipschitz constant $\hat{L}$ to design the shape of the penalisers at all busy locations in the batch may not be optimal. Consider the case where a point in an unexplored region
is still under evaluation. If $\hat{L}$ is large, then the penaliser's radius will be small and we will end up selecting multiple points in the same unexplored region, which is undesirable. 

Therefore, we propose to use a separate Lipschitz constant, which is locally estimated, for each busy location.
Here, ``locality'' is encoded in our choice of kernel and its hyperparameters, e.g. via the lengthscale parameter in the Mat\'{e}rn class of kernels. The use of local Lipschitz constants will enhance the efficiency of exploration because they allow the penaliser to create larger exclusion zones in areas in which we are very uncertain (the surrogate model is near its prior or has low curvature) and smaller penalisation zones in interesting, high-variability, areas. This insight is also corroborated in \cite{blaas19kinky}.

\begin{figure}[t!]
   \centering
    \begin{subfigure}[t]{0.48\linewidth}
        \centering
        \includegraphics[trim={0.5cm 0.0cm 7.7cm 0},clip, width=1.0\linewidth]{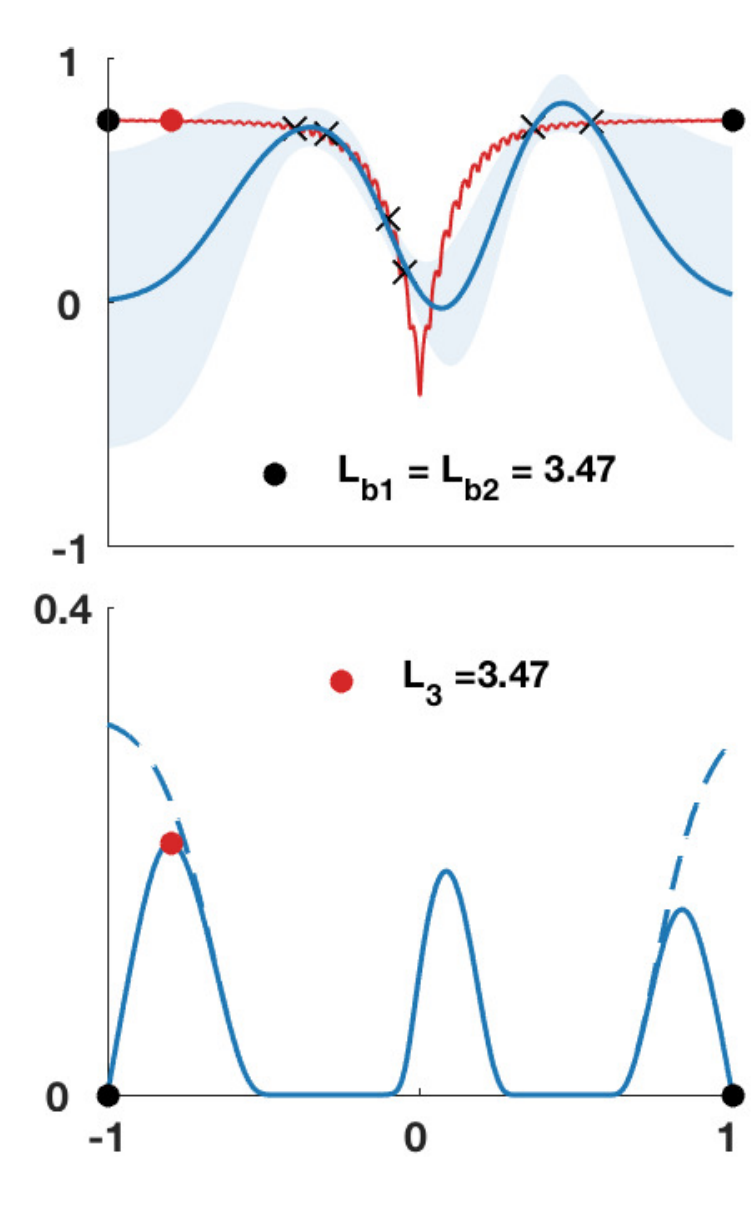}
        \caption{Penalisation with global L} \label{subfig:globalL}
    \end{subfigure}%
    \begin{subfigure}[t]{0.48\linewidth}
        \centering
        \includegraphics[trim={8.2cm 0.0cm 0cm 0},clip, width=1.0\linewidth]{local_lipschitz_effect_async-eps-converted-to.pdf}
        \caption{Penalisation with local L}\label{subfig:localL}
    \end{subfigure}%
    \caption{Different penalisation effects on $\alpha(x)$ of using a single global Lipschitz constant compared to local Lipschitz constants. The top plots in both (\subref{subfig:globalL}) and (\subref{subfig:localL}) show the true objective function (red line), six observed points (black crosses), the GP posterior mean (black line) and variance (blue shade), the two busy locations (black dots) and the next query point (red dot) selected by using the HLP with  global and local Lipschitz constants respectively. The plots in (\subref{subfig:globalL}) show the penalisation effect on busy locations using the same global Lipschitz constant while those in (\subref{subfig:localL}) show the effect of using local Lipschitz constants. It is clear that penalising the busy locations based on local Lipschitz constants allows the algorithm to capture the informative peak at the central region while selection based on the single global Lipschitz constant leads us to revisit the flat region near the boundary due to insufficient penalisation at $x_1$.}
    \label{fig:local_lip_effect}
\end{figure}

We demonstrate the different effects of using approximate global and local Lipschitz constants with a qualitative example. 
In Fig. \ref{subfig:globalL}, the estimated global Lipschitz constant is used for penalisation at both busy locations $x_{b1}=-1$ and $x_{b2}=1$ (denoted as black dots). 
The relatively large value of the global Lipschitz constant ($ \hat{L}_{b1}=  \hat{L}_{b2} =  \hat{L}= 3.47$) due to the high curvature of the surrogate in the central region leads to a small penalisation zone around the two busy locations at the boundary. 
This causes the algorithm to miss the informative region in the centre and instead revisit the region near $x_{b1}$ to choose the new point in the asynchronous batch. 
On the other hand, in Fig. \ref{subfig:localL}, the use of a locally estimated Lipschitz constant allows us to penalise a larger zone around points where the surrogate is relatively flat ($\hat{L}_{b1} =0.712$ for $x_{b1}$), while still penalising smaller regions where there is higher variability ($\hat{L}_{3} =3.45$ at $x_{3}$). 

In our experiments we used a Mat\'{e}rn-52 kernel and defined the local region for evaluating the Lipschitz constant for a batch point $x_j$ to be a hypercube centred on $x_j$ with the length of the each side equal to the lengthscale corresponding to that input dimension.

In summary, we propose a new class of asynchronous BO methods, Penalisation Locally for Asynchronous Bayesian Optimisation Of K workers (PLAyBOOK), which uses analytic penaliser functions to prevent redundant sampling at or near the busy locations in the batch and encourage desirably explorative searching behaviour. 
We differentiate between PLAyBOOK-L, which uses a na\"{i}ve Local penaliser, PLAyBOOK-H, that uses the HLP penaliser, as well as their variations with locally estimated Lipschitz constants, PLAyBOOK-LL and PLAyBOOK-HL. 


\section{Experimental evaluation}
\label{sec:exps}
We begin our empirical investigations by performing a head-to-head comparison of synchronous and asynchronous BO methods, to test the intuitions described in Section \ref{sec:async_vs_synch}. We specifically look at optimisation performance for asynchronous and synchronous variants of the parallel BO methods measured over time and number of evaluations, and we show empirically that asynchronous is preferable over synchronous BO on both counts.

We then experiment with our proposed asynchronous methods (PLAyBOOK-L, PLAyBOOK-H, PLAyBOOK-LL and PLAyBOOK-HL) on a number of benchmark test functions as well as a real-world expensive optimisation task. Our methods are compared against the state-of-the-art asynchronous BO methods, Thompson sampling (TS) \cite{kandasamy2018parallelised}, as well as the Kriging Believer heuristic method (KB) \cite{ginsbourger2010kriging} applied asynchronously. 

For all the benchmark functions, we measure the log of the simple regret $R$, which is the difference between the true minimum value \(f(x^*)\) and the best value found by the BO method:
\begin{equation} \label{eq:simple_regret}
    \log(R) = \log\left \vert f(x^*) - \min_{i=1,\dots, n} f(x_i) \right \vert.
\end{equation}

\subsection{Implementation details}
\label{subsec:implementation}
To ensure a fair comparison, we implemented all methods in Python using the same packages\footnote{
Implementation available at \url{https://github.com/a5a/asynchronous-BO}}.

\begin{figure*}[htb!]
    \centering
        \begin{subfigure}[b]{0.24\textwidth}
        \includegraphics[trim={1cm 1cm 1cm 1cm}, clip, width=\textwidth]{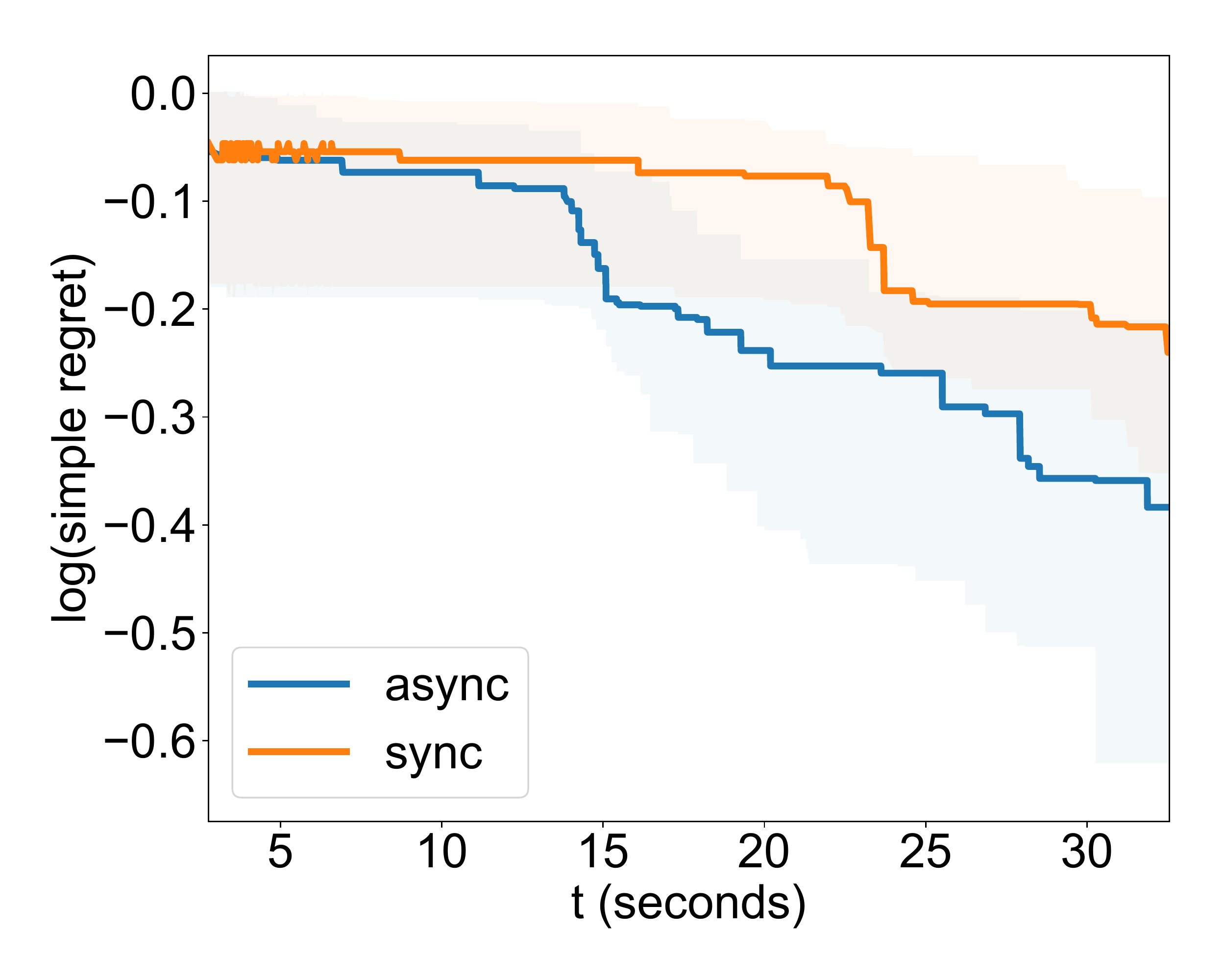}
        \caption{TS, $k = 4$}
        \label{fig:sva_time_ts_4}
    \end{subfigure}
    \begin{subfigure}[b]{0.24\textwidth}
        \includegraphics[width=\textwidth]{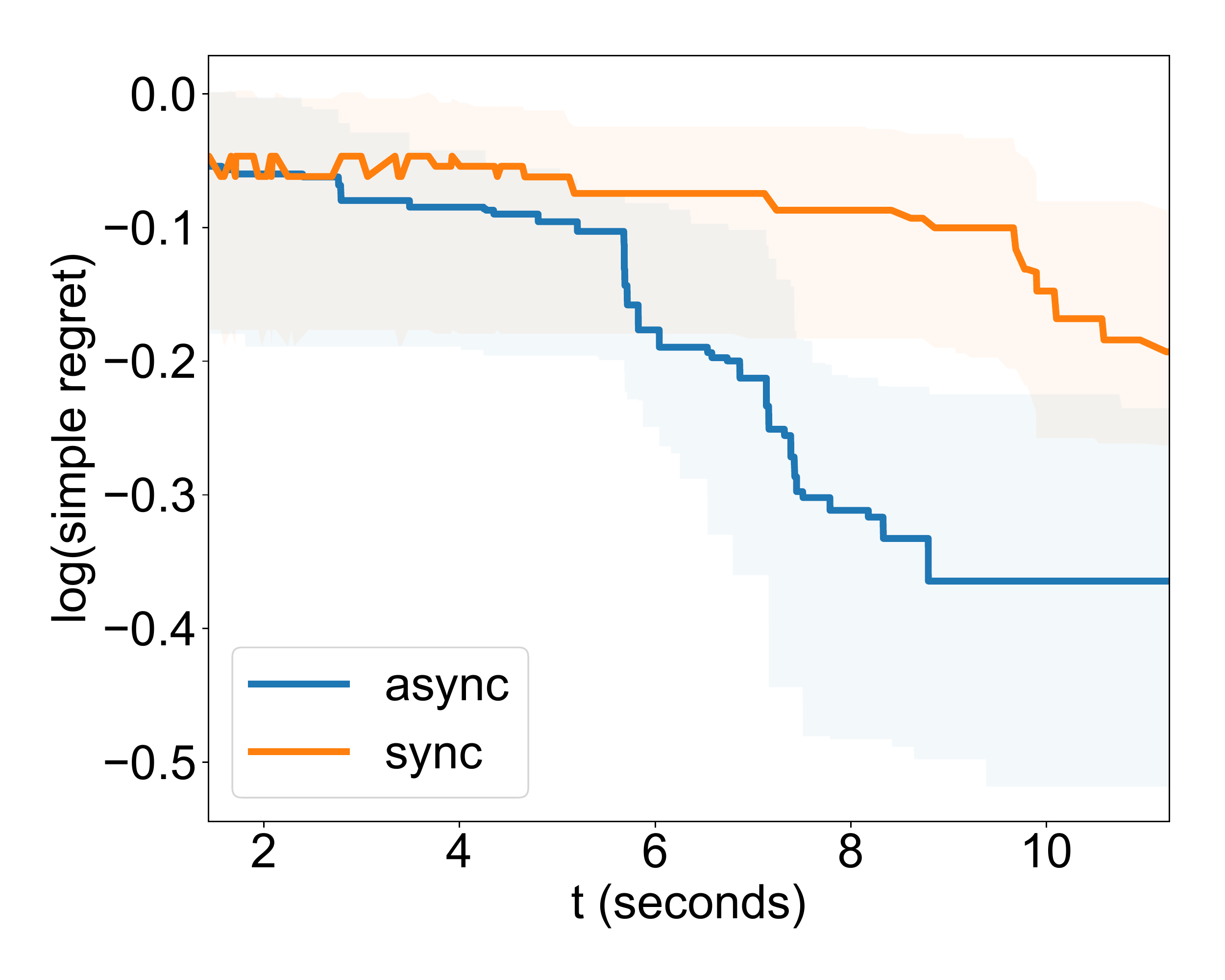}
        \caption{TS, $k = 16$}
        \label{fig:sva_time_ts_16}
    \end{subfigure}
    \begin{subfigure}[b]{0.24\textwidth}
        \includegraphics[width=\textwidth]{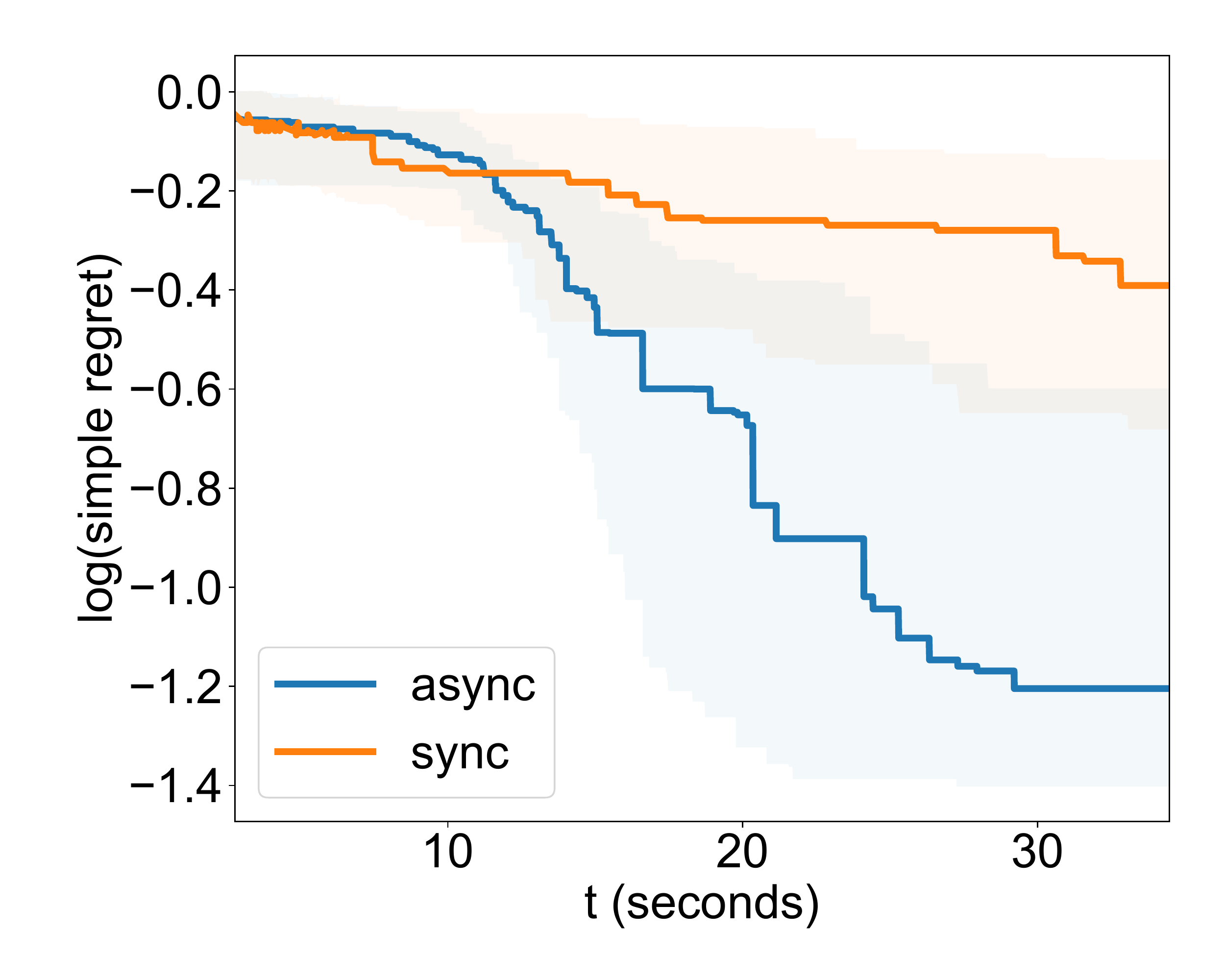}
        \caption{PLAyBOOK-HL, $k = 4$}
        \label{fig:sva_time_hllp_4}
    \end{subfigure}
    \begin{subfigure}[b]{0.24\textwidth}
        \includegraphics[width=\textwidth]{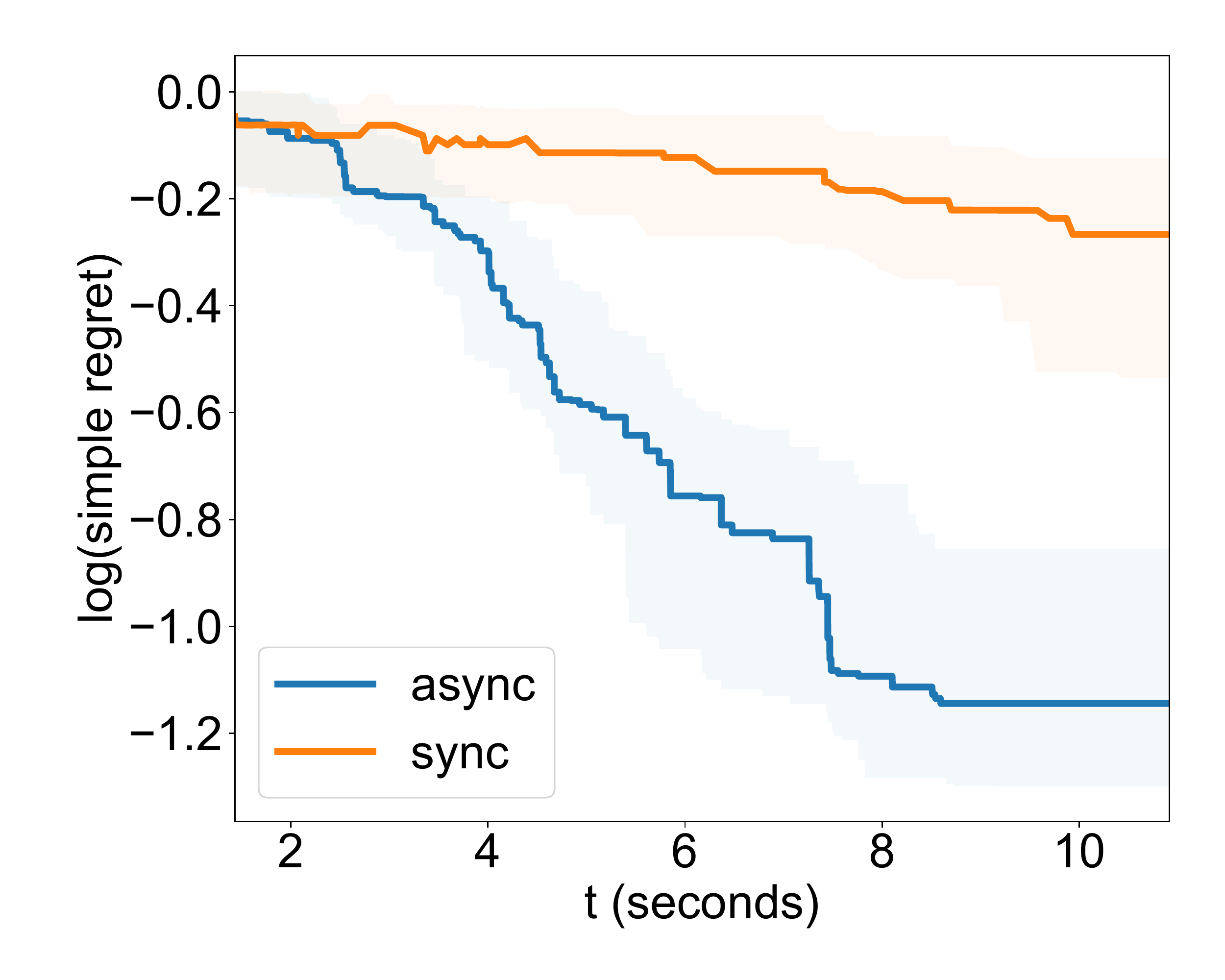}
        \caption{PLAyBOOK-HL, $k = 16$}
        \label{fig:sva_time_hllp_16}
    \end{subfigure}
    
    \begin{subfigure}[b]{0.24\textwidth}
        \includegraphics[trim={1cm 1cm 1cm 1cm}, clip, width=\textwidth]{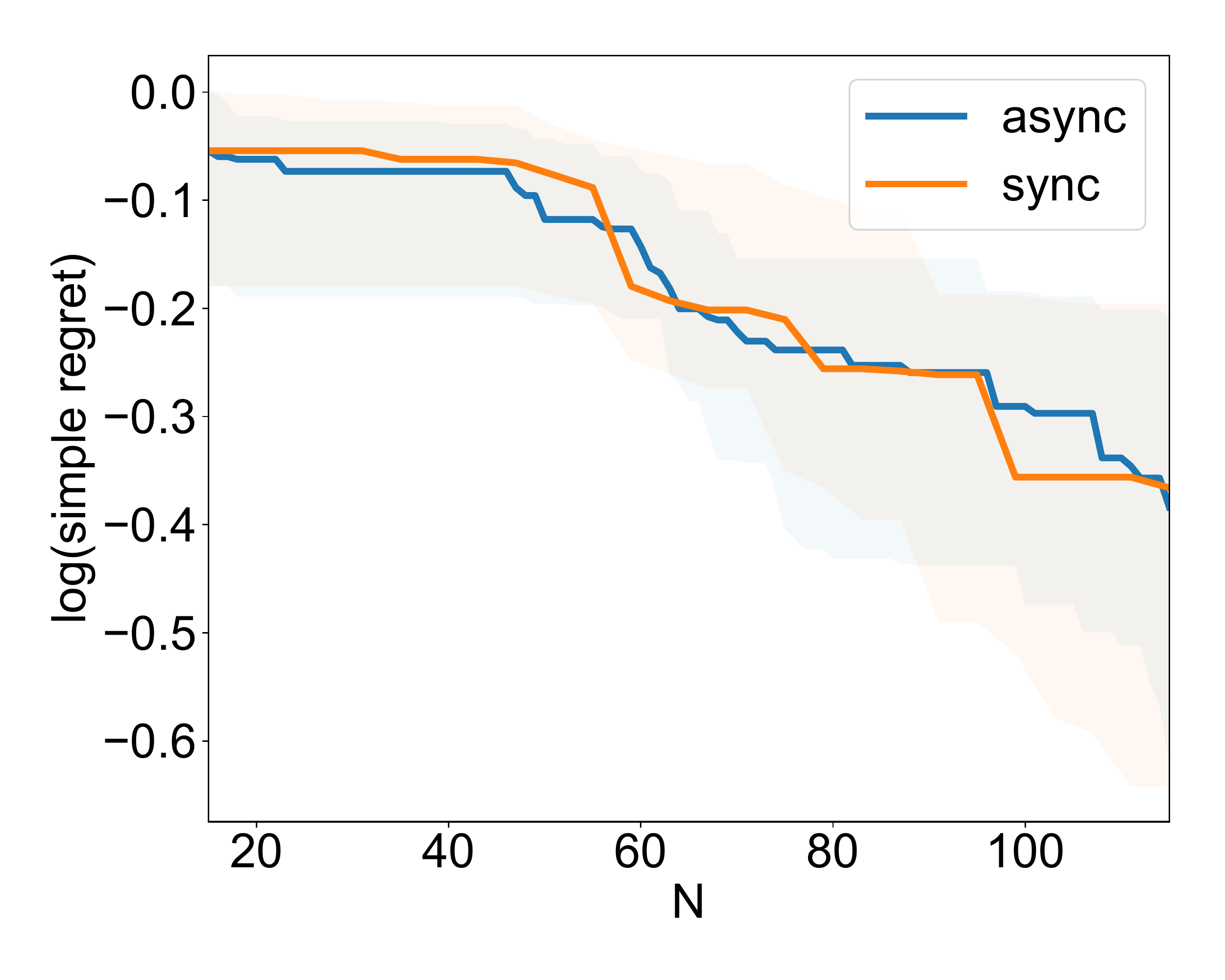}
        \caption{TS, $k = 4$}
        \label{fig:sva_ts_4}
    \end{subfigure}
    \begin{subfigure}[b]{0.24\textwidth}
        \includegraphics[width=\textwidth]{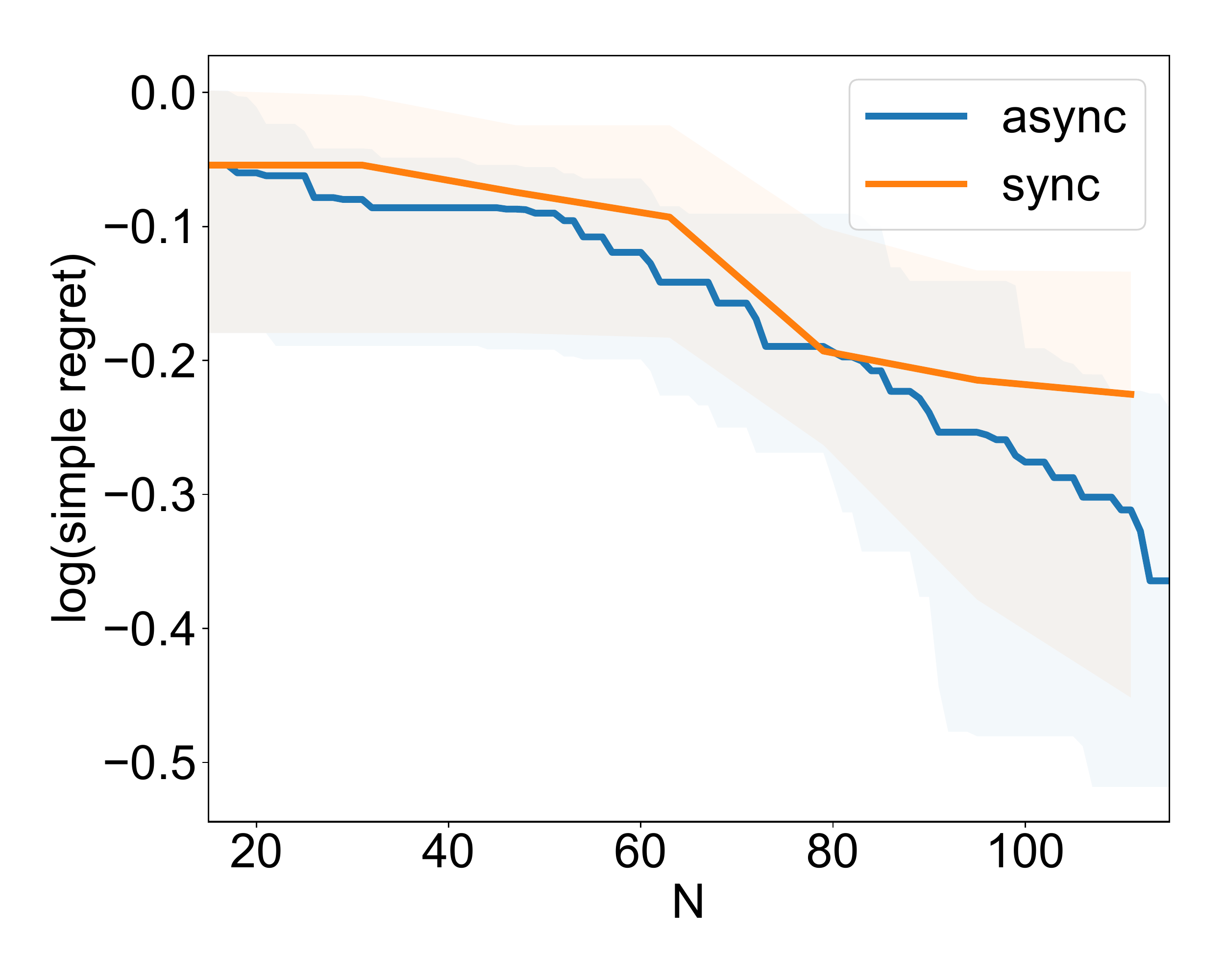}
        \caption{TS, $k = 16$}
        \label{fig:sva_ts_16}
    \end{subfigure}
    \begin{subfigure}[b]{0.24\textwidth}
        \includegraphics[width=\textwidth]{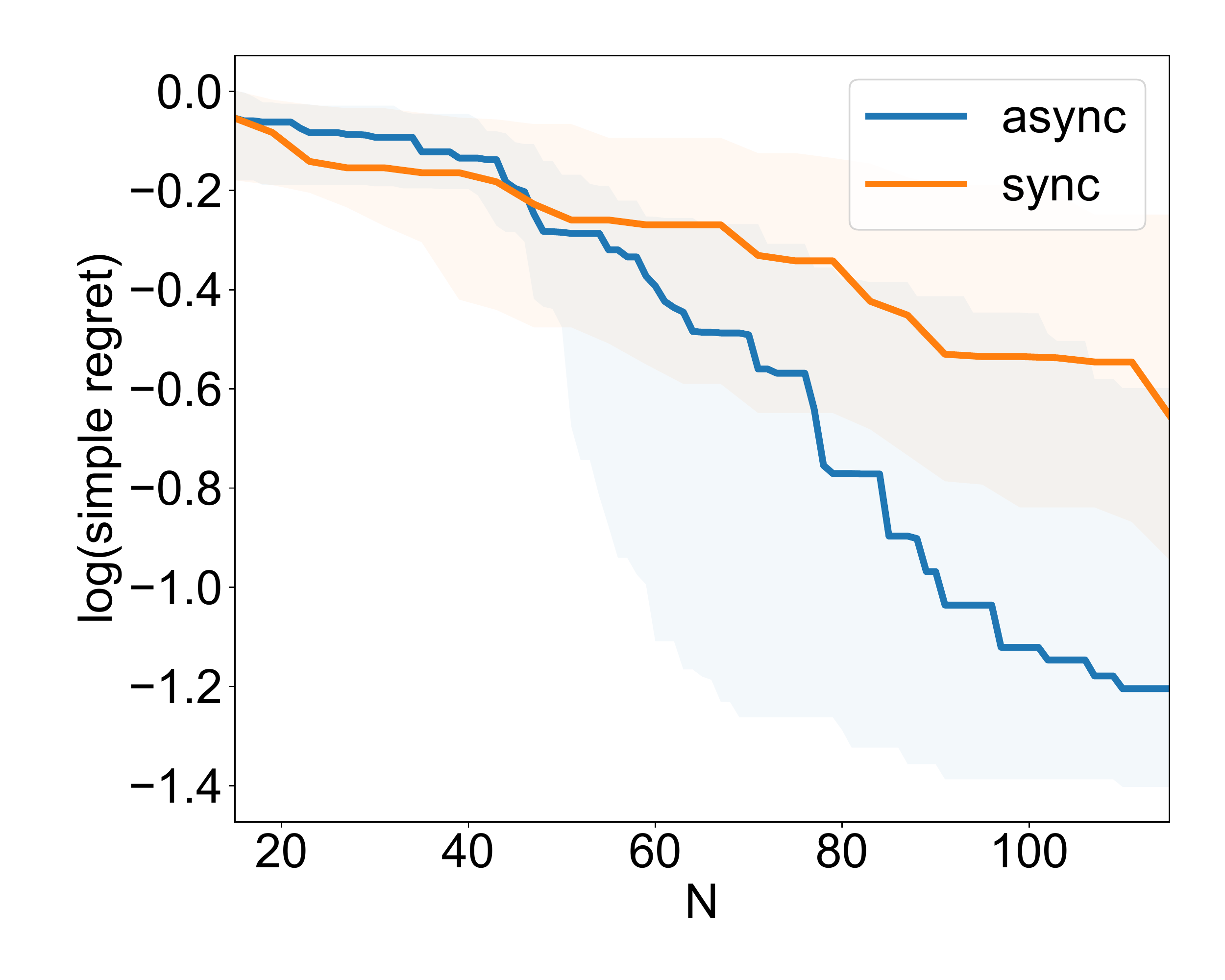}
        \caption{PLAyBOOK-HL, $k = 4$}
        \label{fig:sva_hllp_4}
    \end{subfigure}
    \begin{subfigure}[b]{0.24\textwidth}
        \includegraphics[width=\textwidth]{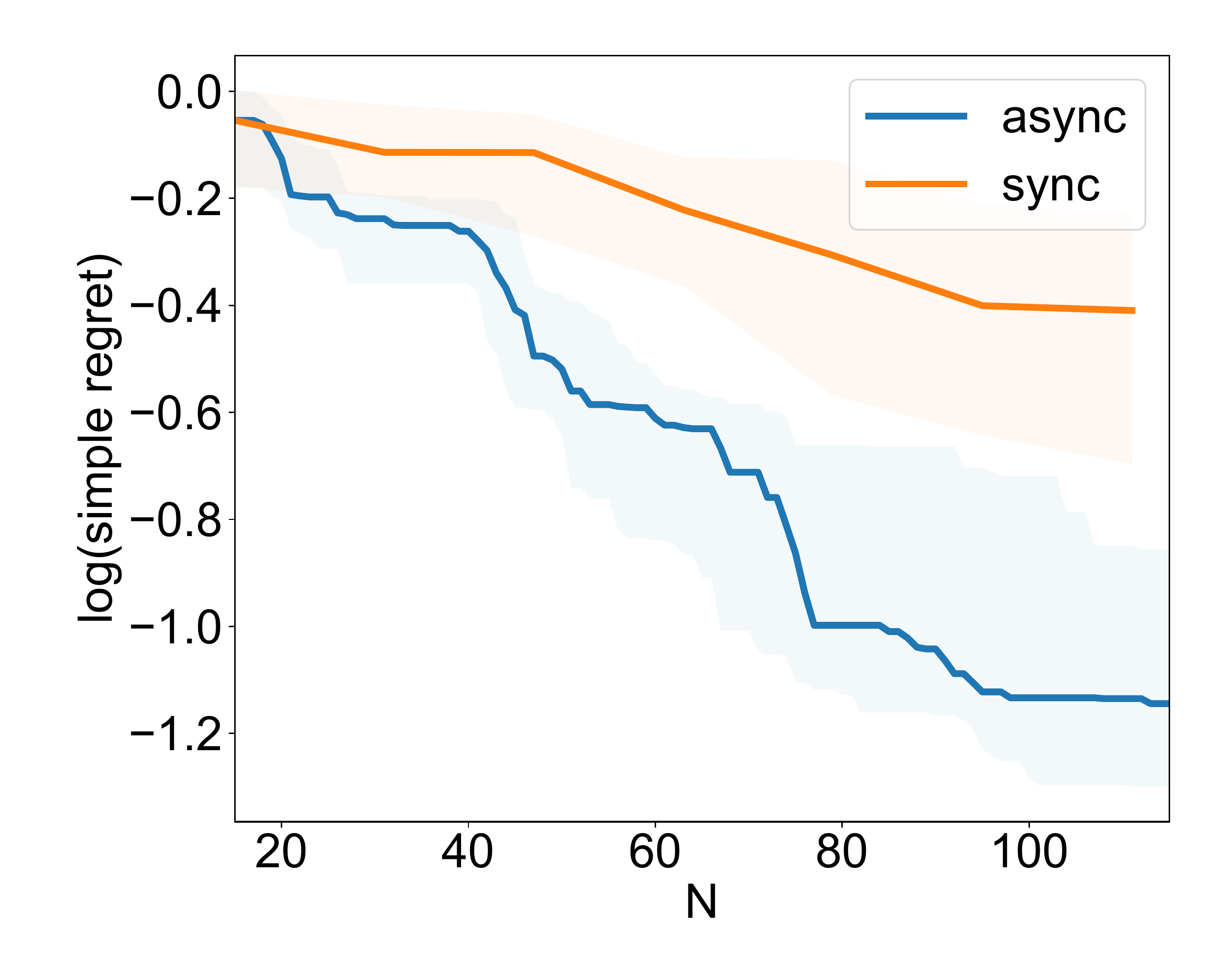}
        \caption{PLAyBOOK-HL, $k = 16$}
        \label{fig:sva_hllp_16}
    \end{subfigure}
    \caption{A head-to-head comparison of synchronous (orange) vs asynchronous (blue) versions of a parallel BO method. The median (solid line) and quartiles (shaded region) of the regret for optimising ack-5D for 30 random initialisations are shown. The top row shows regret vs evaluation time and the bottom row shows regret vs number of evaluations. Notice how asynchronous methods in the top row outperform their synchronous counterparts in terms of evaluation time. Note also how asynchronous methods in the bottom row outperform their synchronous counterparts in terms of sample efficiency, too.
     }
    \label{fig:sva_time_and_n}
\end{figure*}

In all experiments, we used a zero-mean Gaussian process surrogate model with a Mat\'{e}rn-52 kernel with ARD. 
We optimised the kernel and likelihood hyperparameters by maximising the log marginal likelihood.
For the benchmark test functions, we fixed the noise variance to $\sigma^2=10^{-6}$ and started with $3*d$ random initial observations. Each experiment was repeated with $30$ different random initialisations and the input domains for all experiments were scaled to \([-1, 1]^d\).

All methods except TS used UCB as the acquisition function \(\alpha(x)\). For our PLAyBOOK-H and PLAyBOOK-HL, we choose $\gamma=1$ and $p=-5$ in the HLP (Eq. \eqref{eq:hlp}).
For TS, we use 10,000 sample points for each batch point selection. 
For the other methods, we evaluate \(\alpha(x)\) at 3,000 random locations and then choose the best one after locally optimising the best 5 samples for a small number of local optimisation steps.

We evaluate the performance of the different batch BO strategies using popular
global optimisation test functions\footnote{Details for these and other challenging global optimisation test functions can be found at \url{https://www.sfu.ca/~ssurjano/optimization.html}}. We show results for the Eggholder function defined on $\mathbb{R}^2$ (egg-2D), the Ackley function defined on $\mathbb{R}^5$ and the Michalewicz function defined on $\mathbb{R}^{10}$ (mic-10D). Results for experiments on different example tasks can be found in Sections 2 and 3 in the supplementary materials.


\subsection{Synchronous vs asynchronous BO}
\label{subsec:exps-sync-vs-async}

In this section we address the question of choosing between asynchronous and synchronous BO. In order to investigate their relative merits, we compared asynchronous and synchronous BO methods' performance as a function of wall-clock time and number of evaluations.

\subsubsection{Evaluation time}
In order to facilitate this comparison, we needed to inject a measure of runtime for different tasks, as the test functions can be evaluated instantaneously. 
We followed the procedure proposed in \citet{kandasamy2018parallelised} to sample an evaluation time for each task so as to simulate the asynchronous setting. 
We chose to use a half-normal distribution with scale parameter \(\sigma = \sqrt{\pi/2}\), which gives us a distribution of runtime values with mean at 1. 

Results on the ack-5D task are shown in Figs. \ref{fig:sva_time_and_n}(\subref{fig:sva_time_ts_4})-(\subref{fig:sva_time_hllp_16}). Due to space constraints, results for further experiments can be found in Section 2 of the supplementary materials.
We know that asynchronous BO has the advantage over synchronous BO in terms of utilisation of resources, as shown qualitatively in Fig. \ref{fig:sync_vs_async}, simply due to the fact that any available worker is not required to wait for all evaluations in the batch to finish before moving on. Therefore, given the same time budget, a greater number of evaluations can be taken in the asynchronous setting than in the synchronous setting, which, as confirmed by our experiments, translates to faster optimisation of $f$ in terms of the total (wall-clock) time spent evaluating tasks.

\begin{figure*}[htb!]
    \centering
    \begin{subfigure}[b]{0.3\textwidth}
        \includegraphics[
        width=\textwidth]{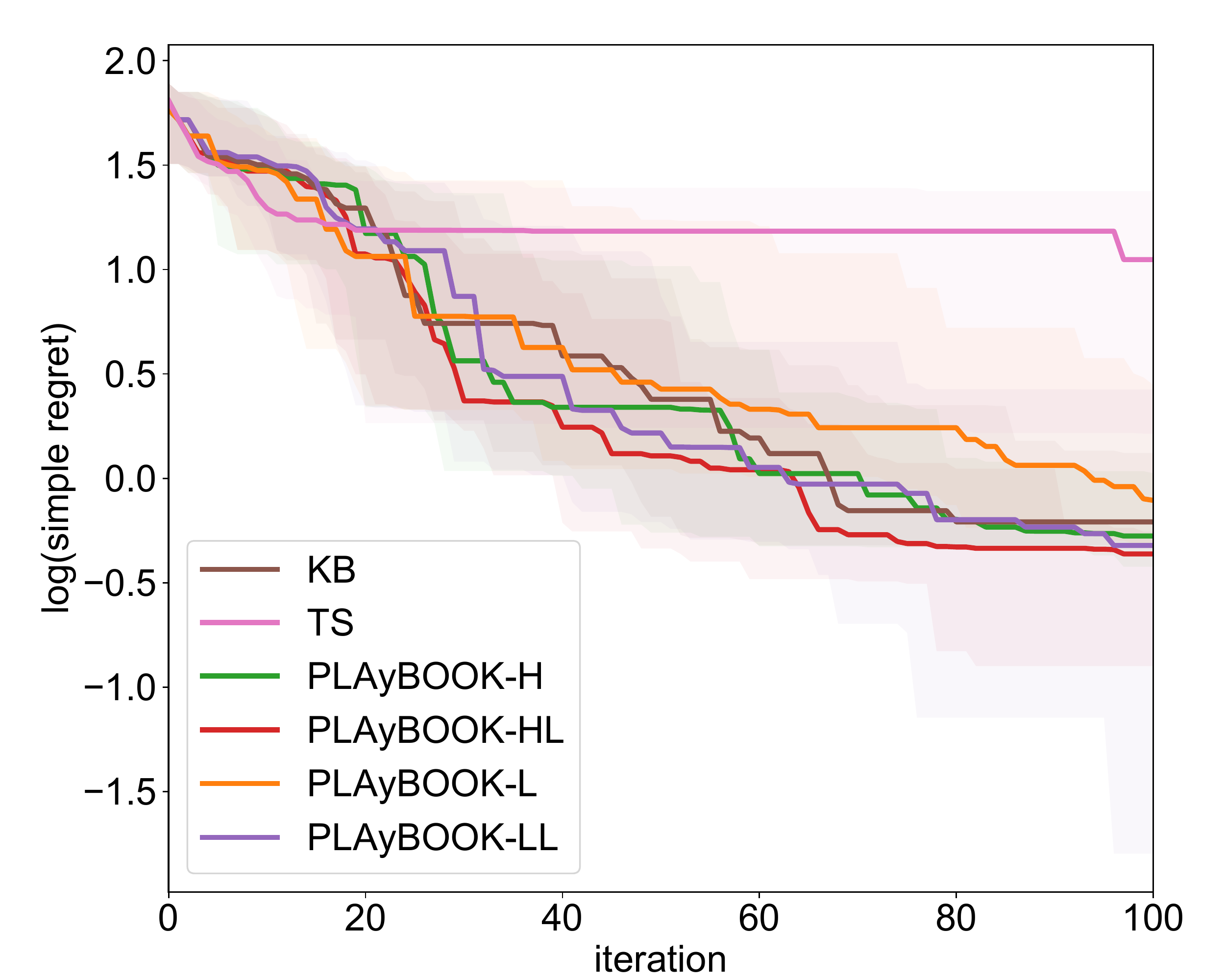}
        \caption{egg-2d, $k=4$}
        \label{fig:egg2_bs4}
    \end{subfigure}
    \begin{subfigure}[b]{0.3\textwidth}
        \includegraphics[width=\textwidth]{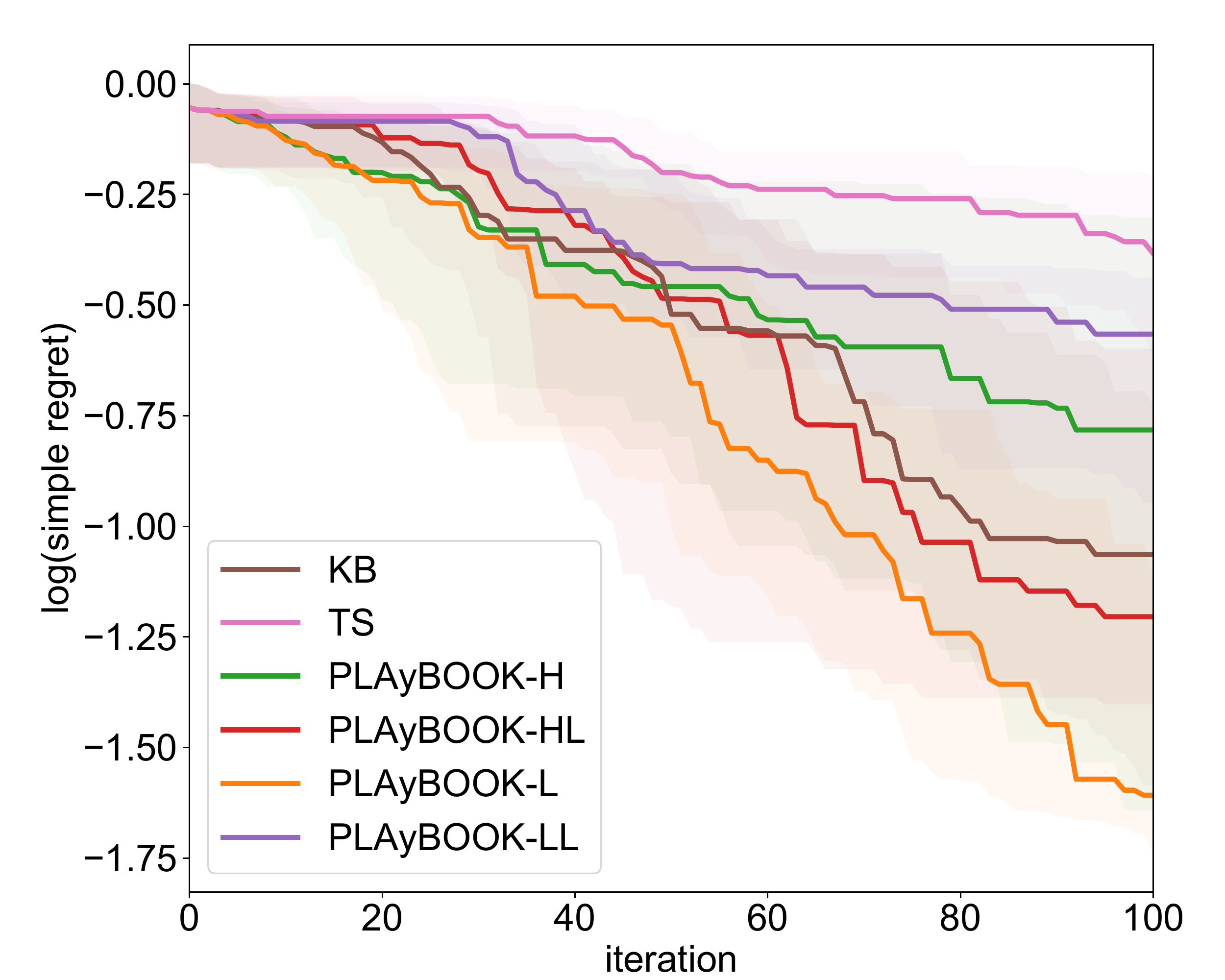}
        \caption{ack-5d, $k=4$}
        \label{fig:ack5_bs4}
    \end{subfigure}
    \begin{subfigure}[b]{0.3\textwidth}
        \includegraphics[width=\textwidth]{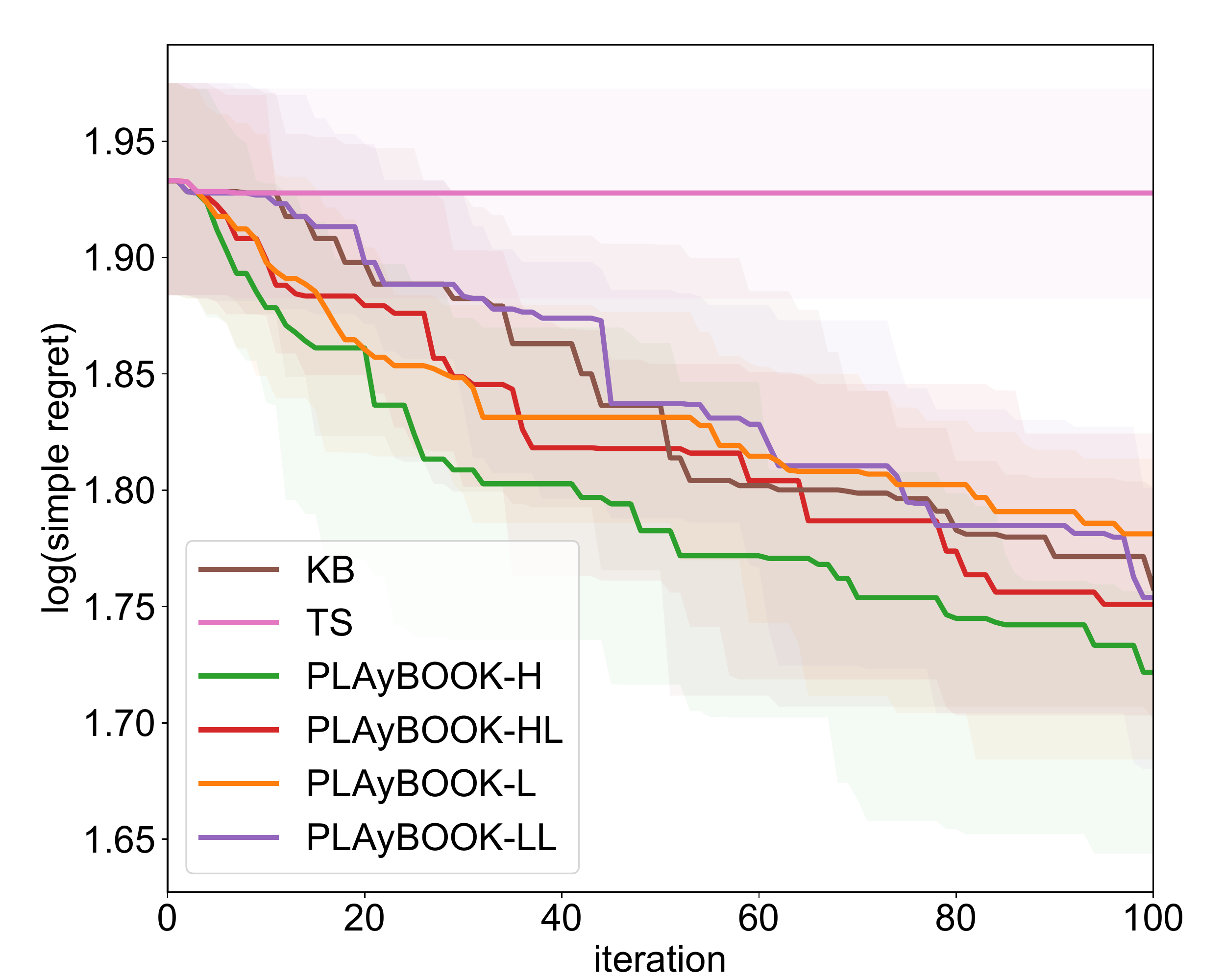}
        \caption{mic-10D, $k=4$}
        \label{fig:mi5_bs4}
    \end{subfigure}
    
    \begin{subfigure}[b]{0.3\textwidth}
        \includegraphics[width=\textwidth]{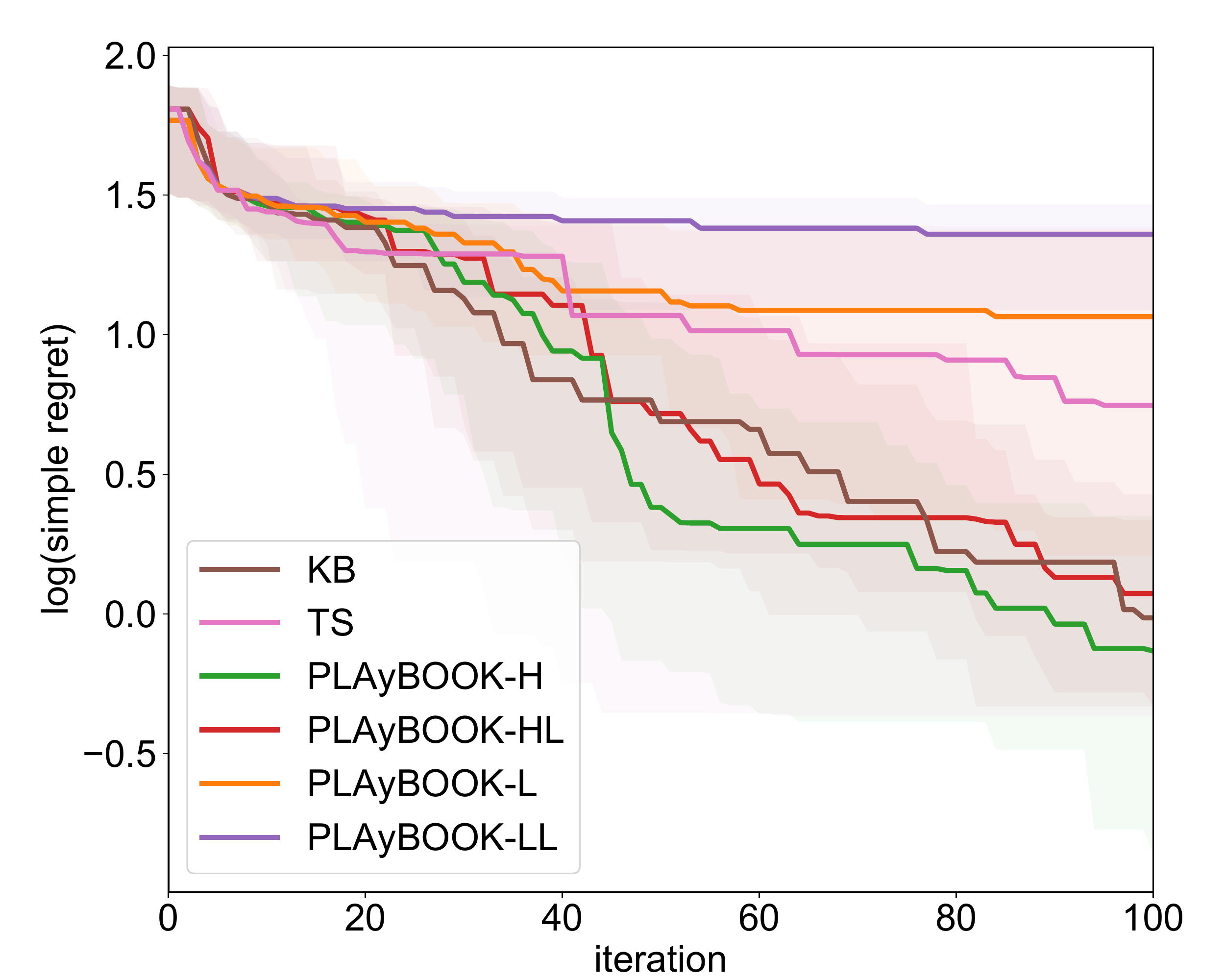}
        \caption{egg-2d, $k=16$}
        \label{fig:egg2_bs16}
    \end{subfigure}
    \begin{subfigure}[b]{0.3\textwidth}
        \includegraphics[width=\textwidth]{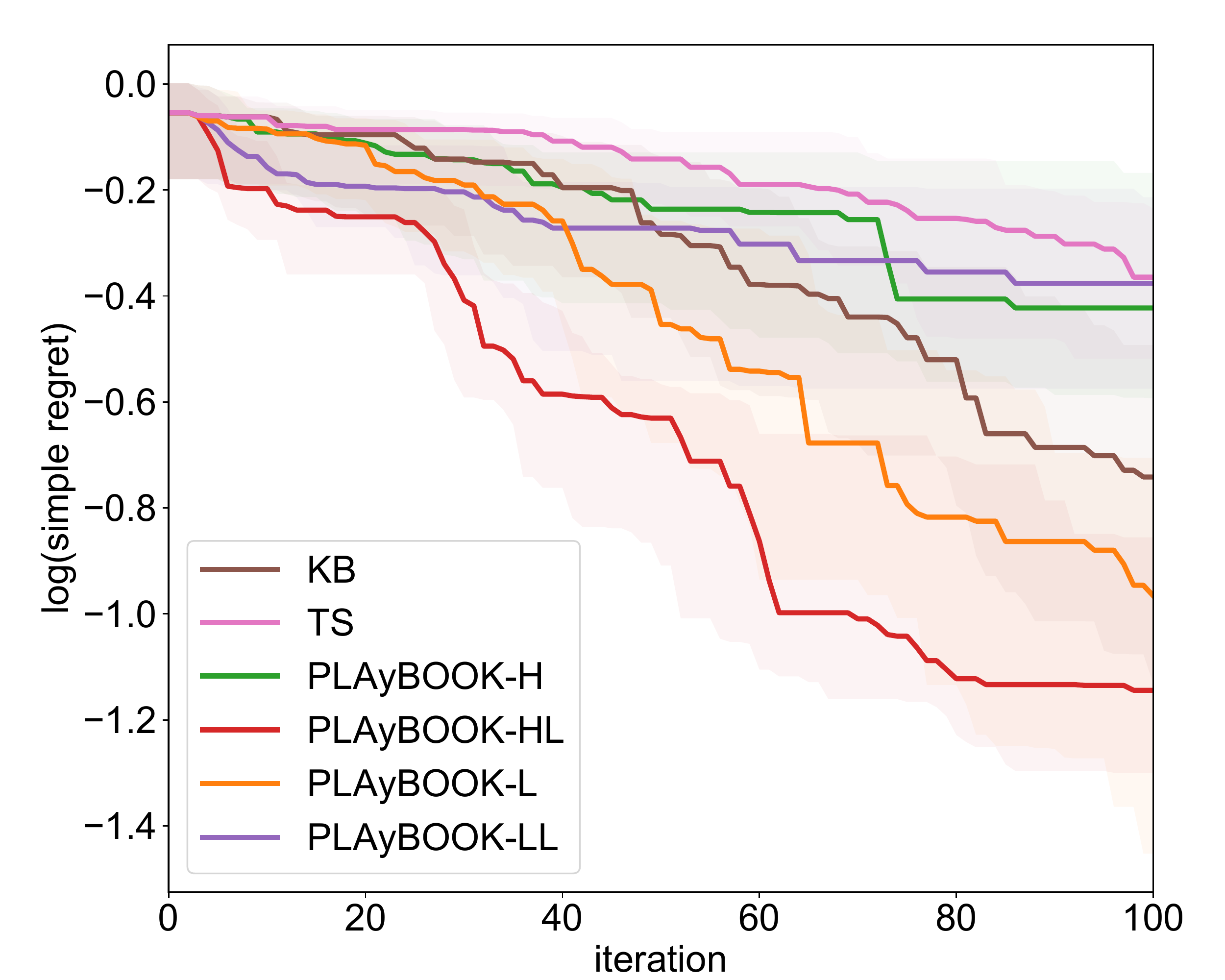}
        \caption{ack-5d, $k=16$}
        \label{fig:ack5_bs16}
    \end{subfigure}
    ~ 
    \begin{subfigure}[b]{0.3\textwidth}
        \includegraphics[width=\textwidth]{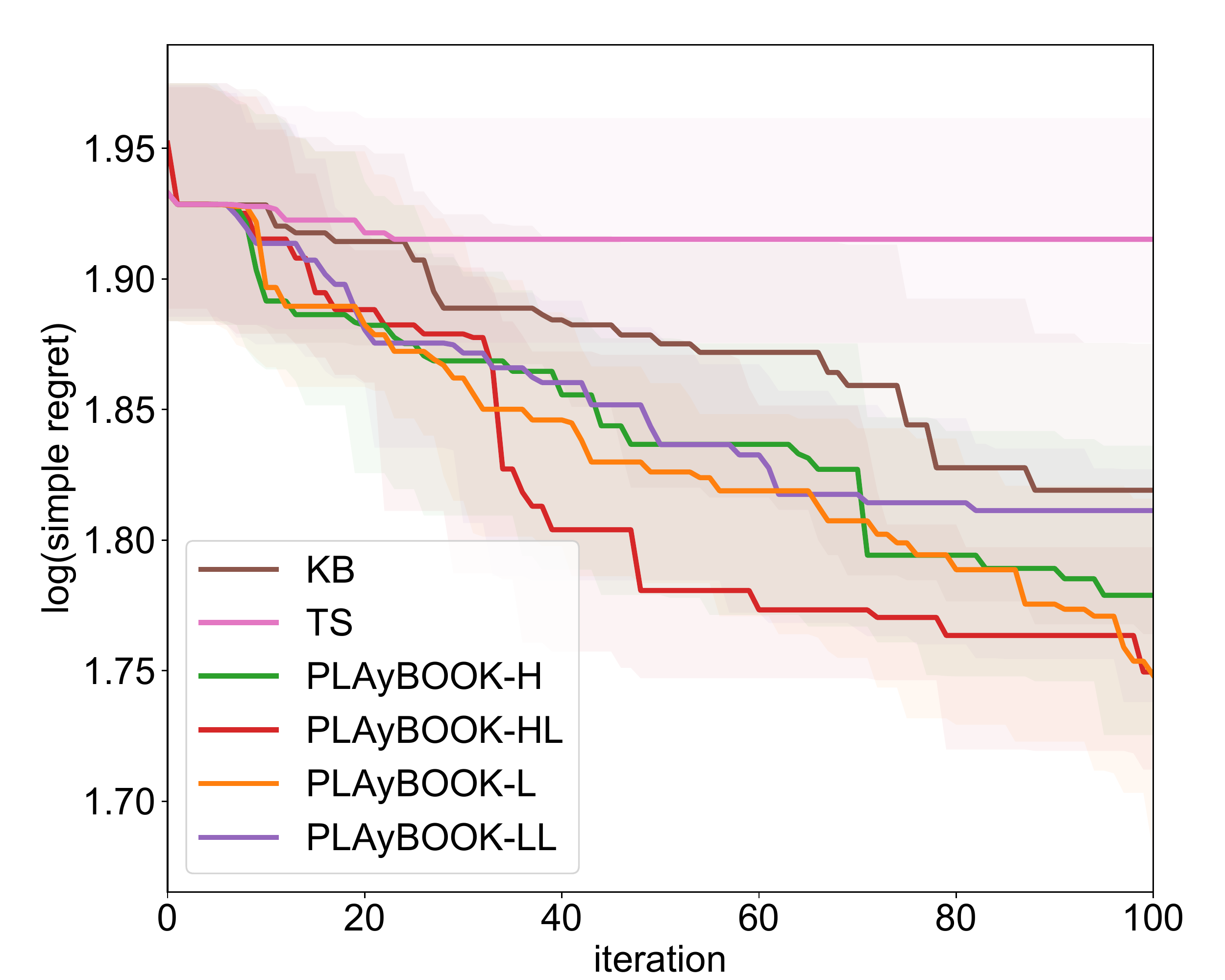}
        \caption{mic-10D, $k=16$}
        \label{fig:mi5_bs16}
    \end{subfigure}
    \caption{The median (solid line) and quartiles (shaded region) of the regret for different asynchronous BO methods on the global optimisation test functions for 30 random intialisations is shown. We can see that our proposed PLAyBOOK methods perform competitively, especially when we start choosing larger batch sizes.}
    \label{fig:async-bo-on-math-funcs}
\end{figure*}

\subsubsection{Number of evaluations}
A more interesting question to answer is whether asynchronous BO methods are really less data efficient than synchronous BO methods as discussed in Section \ref{sec:async_vs_synch}. Figs. \ref{fig:sva_time_and_n}(\subref{fig:sva_ts_4})-(\subref{fig:sva_hllp_16}) show a subset of the experiments we conducted. More results can be found in Section 2 in the supplementary materials.

An unexpected yet interesting behaviour we note is that as $k$ increases, the PLAyBOOK methods tend to clearly outperform their respective synchronous counterparts even in terms of sample efficiency. This observation runs counter to the guidance provided in \citet{kandasamy2018parallelised} and such behaviour is less evident for the other two batch methods, TS and KB.

We think this observation may be explained by the difference in nature between the PLAyBOOK and TS/KB: in the case of TS we rely on stochasticity in sampling, and in KB we are re-computing the posterior variance and \(\alpha(x)\) each time a batch point is selected. 
The penalisation-based methods, on the other hand, simply down-weight the acquisition function, and in the synchronous case these penalisers coincide with the high-value regions of the acquisition function. 
This means that unless the acquisition function has a large number of spaced-out peaks, we will quickly be left without high-utility locations to choose new batch points from. 

This seems to be the reason for the superior performance of asynchronous PLAyBOOK methods over their synchronous variants because they benefit from the fact that the busy locations being penalised do not necessarily coincide with the peaks in \(\alpha(x)\), as the surrogate used to compute \(\alpha(x)\) is more informed than the one used to decide the locations of the busy locations previously. This means that points with high utilities are more likely to to be preserved.

Taking into account the fact that the asynchronous PLAyBOOK methods tend to perform at least equally well, if not significantly better than their synchronous variants on both time and efficiency, and that the asynchronous PLAyBOOK methods gain more advantage over synchronous ones as the batch size increases, we believe that this points to the fact that penalisation-based methods are inherently better suited as asynchronous methods. Hence, for users that are running parallel BO and have selected LP, we recommend they consider running PLAyBOOK instead due to its attractive benefits.


\subsection{Asynchronous parallel BO}
\label{subsec:exps-lp-async}

Now that we have strengthened the appeal of asynchronous BO, we turn to evaluating PLAyBOOK against existing asynchronous BO methods. 

\subsubsection{Synthetic experiments}
We ran PLAyBOOK and competing asynchronous BO methods on the global optimisation test functions described in Section \ref{subsec:implementation}. The results are shown in Fig. \ref{fig:async-bo-on-math-funcs}, and more results on different optimisation problems are provided in Section 3 in the supplementary materials.

On the global optimisation test functions we noted that in most cases PLAyBOOK outperforms the alternative asynchronous methods TS and KB. 
The TS algorithm performed poorly on this test, which we believe is caused by the fact that TS relies heavily on the surrogate's uncertainty to explore new regions. 

PLAyBOOK methods show strong performance, achieving better optimisation performance than both TS and KB baselines.

\subsubsection{Real-world optimisation}

\begin{figure}[htb!]
   \centering
    \begin{subfigure}[t]{0.48\linewidth}
        \centering
        \includegraphics[trim={0.1cm 0.1cm 0.0cm 0},clip, width=1.0\linewidth]{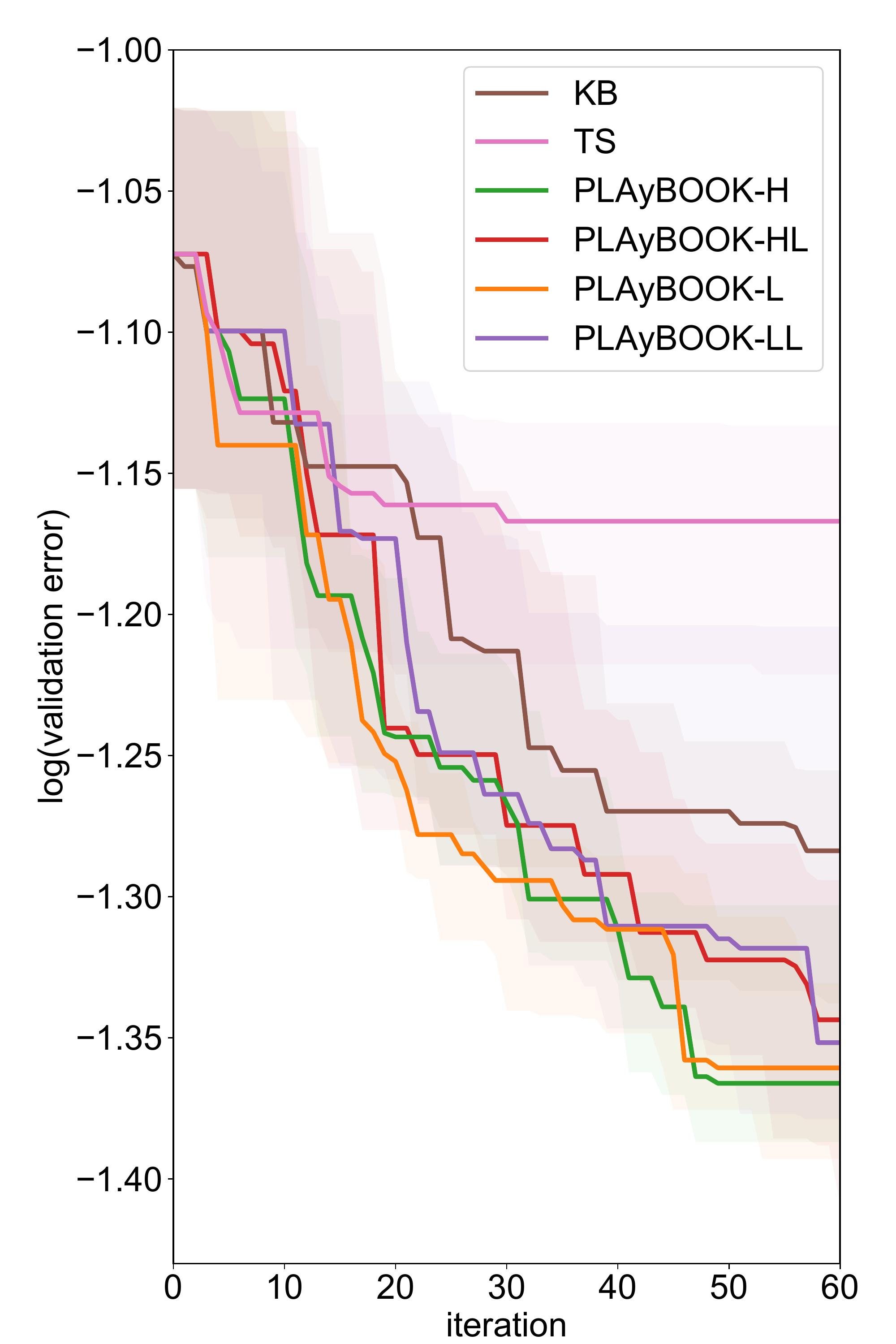}
        \caption{$k=2$} \label{subfig:cnn6_cifar10_bs2}
    \end{subfigure}%
    \begin{subfigure}[t]{0.48\linewidth}
        \centering
        \includegraphics[trim={0.1cm 0.1cm 0cm 0},clip, width=1.0\linewidth]{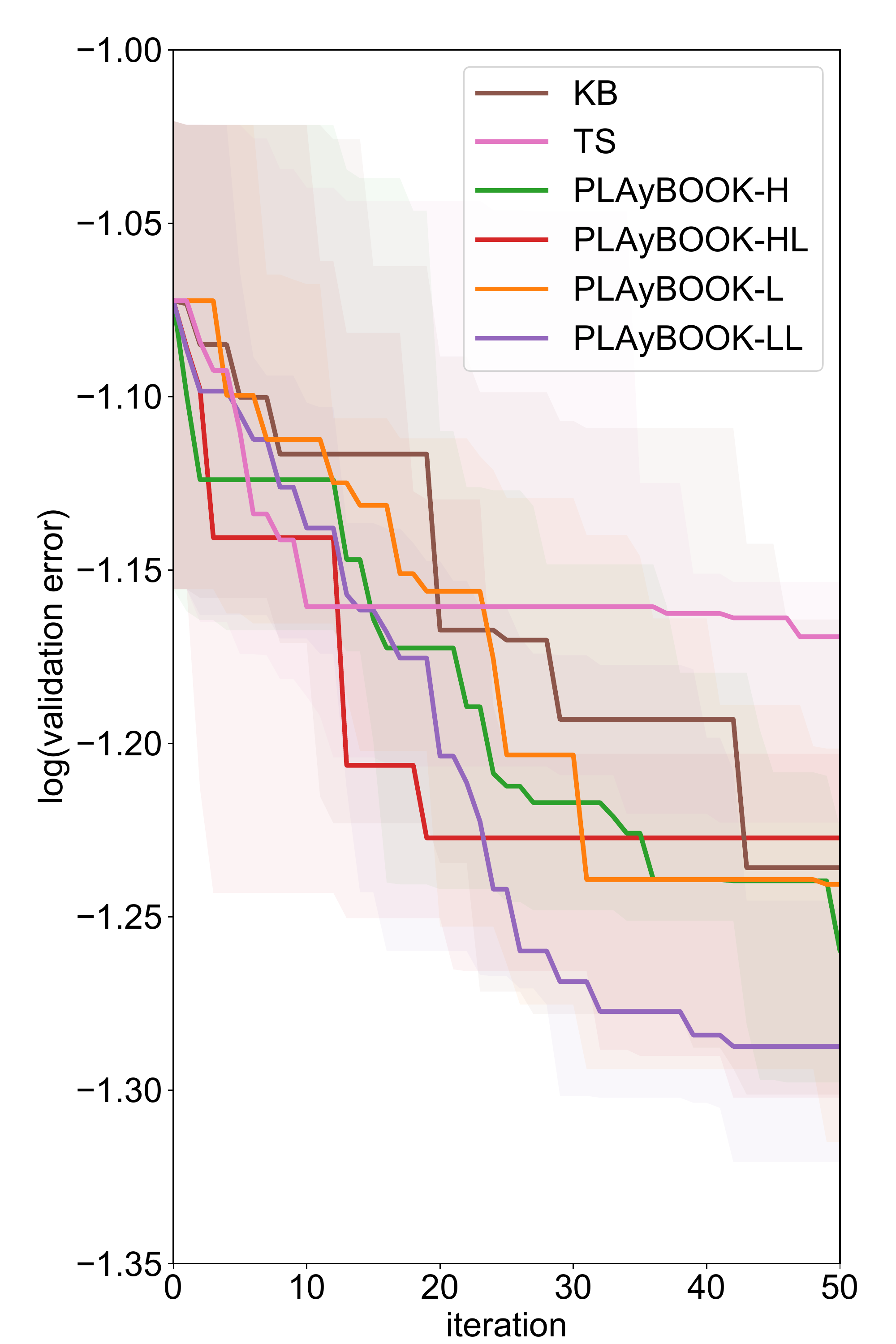}
        \caption{$k=4$}\label{subfig:cnn6_cifar10_bs4}
    \end{subfigure}%
    \caption{Asynchronous optimisation of $9$ hyperparameters of a $6$-layer CNN for image classification on the CIFAR10 dataset. The network is trained on half of the training set and evaluated on the second half. The objective being minimised is the classification accuracy on the validation set. PLAyBOOK outperforms both KB and TS in this expensive optimisation task.}
    \label{fig:cnn6_cifar}
\end{figure}

We further experimented on a real-world application of tuning the hyperparameters of a 6-layer Convolutional Neural Network (CNN)\footnote{Follow the implementation in \url{https://blog.plon.io/tutorials/cifar-10-classification-using-keras-tutorial/}} for an image classification task on CIFAR10 dataset \cite{krizhevsky2009learning}. The 9 hyperparameters that we optimise with BO are the learning rate and momentum for the stochastic gradient descent training algorithm, the batch size used and the number of filters in each of the six convolutional layers. We trained the CNN on half of the training set for 20 epochs and each function evaluation returns the validation error of the model. We tested the use of $k=2$ and $k=4$ parallel workers to run this real-world experiment. The results are shown in Figs. \ref{subfig:cnn6_cifar10_bs2} and \ref {subfig:cnn6_cifar10_bs4}. 

We can see that for both $k=2$ and $k=4$ parallel settings, all PLAyBOOK methods outperform the other asynchronous methods, TS and KB. In the case of $k=2$ ($2$ parallel processors), only one busy location is penalised in each batch so there is little gain from using a locally estimated Lipschitz constant. However, as the batch size increases to $k=4$, we see that methods using estimated Lipschitz constants (PLAyBOOK-LL and PLAyBOOK-HL) show faster decrease in validation error than PLAyBOOK-L and PLAyBOOK-H with PLAyBOOK-LL demonstrating the best performance.

We then took the final configurations recommended by each asynchronous BO method in the $k=2$ and $k=4$ settings and retrained the CNN model on the full training set of 50K images for 80 epochs. The accuracy on the test set of 10K images achieved with the best model chosen by each BO method is shown in Table \ref{table:full_cnn_test}. In both settings, our PLAyBOOK methods  achieve superior performance over TS with PLAyBOOK-H providing the best test accuracy when $k=2$ and PLAyBOOK-LL doing the best when $k=4$.

\begin{table}[htb!]
\caption{Test accuracy (\%) on CIFAR-10 after training the best model chosen by  various asynchronous BO methods for $80$ epochs} \label{table:full_cnn_test}
\begin{center} 
\begin{tabular}{ccccccc}
\toprule
\multicolumn{1}{c}{\multirow{2}{*}{$k$}} & \multirow{2}{*}{{TS}} & \multirow{2}{*}{{KB}} & \multicolumn{4}{c}{{PLAyBOOK}}\\ \cmidrule{4-7} 
              &      &       & {L} & {H} & {LL}             & {HL} \\ \midrule
 2       & 81.0  & 83.9 & 84.7      & \textbf{85.2}  & 84.1 &  84.9 \\ 
 4       & 81.2  & 82.8 & 82.5      & 83.8       & \textbf{84.2} &  83.0    \\
\bottomrule
\end{tabular}
\end{center}
\end{table}



\section{Conclusions}
We argue for the use of asynchronous (over synchronous) Bayesian optimisation (BO), and provide supporting empirical evidence.  Additionally, we developed a new approach, PLAyBOOK, for asynchronous BO, based on penalisation of the acquisition function using information about tasks that are still under evaluation. Empirical evaluation on synthetic functions and a real-world optimisation task showed that PLAyBOOK improves upon the state of the art. Finally, we demonstrate that, for penalisation-based batch BO, PLAyBOOK's asynchronous BO is more efficient than synchronous BO in both wall-clock time and the number of samples.

\section*{Acknowledgements}
We thank our colleagues at the Machine Learning Research Group at the University of Oxford, especially Edward Wagstaff, for providing discussions that assisted the research. We would also like to express our gratitude to our anonymous reviewers for their valuable comments and feedback.

Computational resources were supported by Arcus HTC and JADE HPC at the University of Oxford and Hartree national computing facilities, UK.

\bibliography{asyncBO}
\bibliographystyle{icml2019}

\input{supplementary.tex}

\end{document}

%% file: supplementary.tex







\twocolumn[
\icmltitle{Supplementary material}

\vskip 0.3in
]

\setcounter{section}{0}
\section{Summary of experimental tasks}
As mentioned in the main text, we conducted empirical evaluations on a large number of synthetic test problems:
\begin{itemize}
    \item The tasks mat-2 and mat-6 refer to functions drawn from a Gaussian process(GP) with Mat\'{e}rn-52 kernel in $\mathbb{R}^{2}$ and $\mathbb{R}^{6}$ respectively. 
    \item The global optimisation tasks\footnote{Details for these and other challenging global optimisation test functions can be found at \url{https://www.sfu.ca/~ssurjano/optimization.html}} that we considered are the Ackley function defined on $\mathbb{R}^5$ and $\mathbb{R}^{10}$ (ack-5 and ack-10), the Michalewicz function defined on $\mathbb{R}^{5}$ and $\mathbb{R}^{10}$ (mic-5 and mic-10) and the Eggholder function in $\mathbb{R}^{2}$ (egg-2).
    \item We also selected a robot pushing simulation experiment, which was first explored in a BO context by \citet{wang2017maxvalue}. Here the task is to learn the correct pushing action to minimise the distance of the robot to a goal. The problem has 4 inputs: the robot's location \((r_x, r_y)\), the angle of the pushing force \(r_\theta\) and the pushing duration \(t_r\). We used the input space suggested by \citet{wang2017maxvalue}.
\end{itemize}

\section{Asynchronous vs. synchronous parallel BO}
\begin{figure*}[htb!]
    \centering
        \begin{subfigure}[b]{0.24\textwidth}
        \includegraphics[trim={1cm 1cm 1cm 1cm}, clip, width=\textwidth]{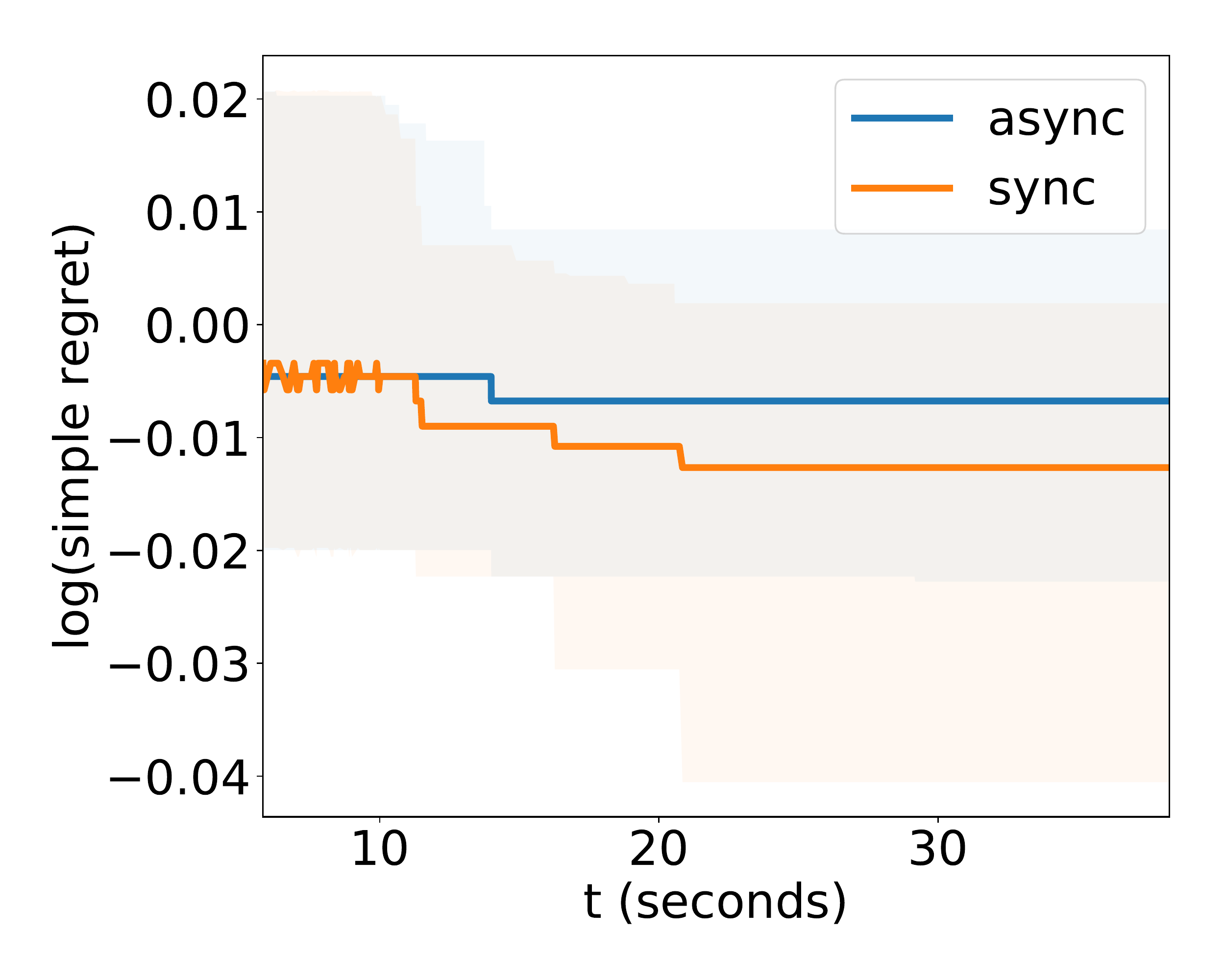}
        \caption{TS, $k = 4$}
        \label{fig:sva_time_ts_4}
    \end{subfigure}
    \begin{subfigure}[b]{0.24\textwidth}
        \includegraphics[width=\textwidth]{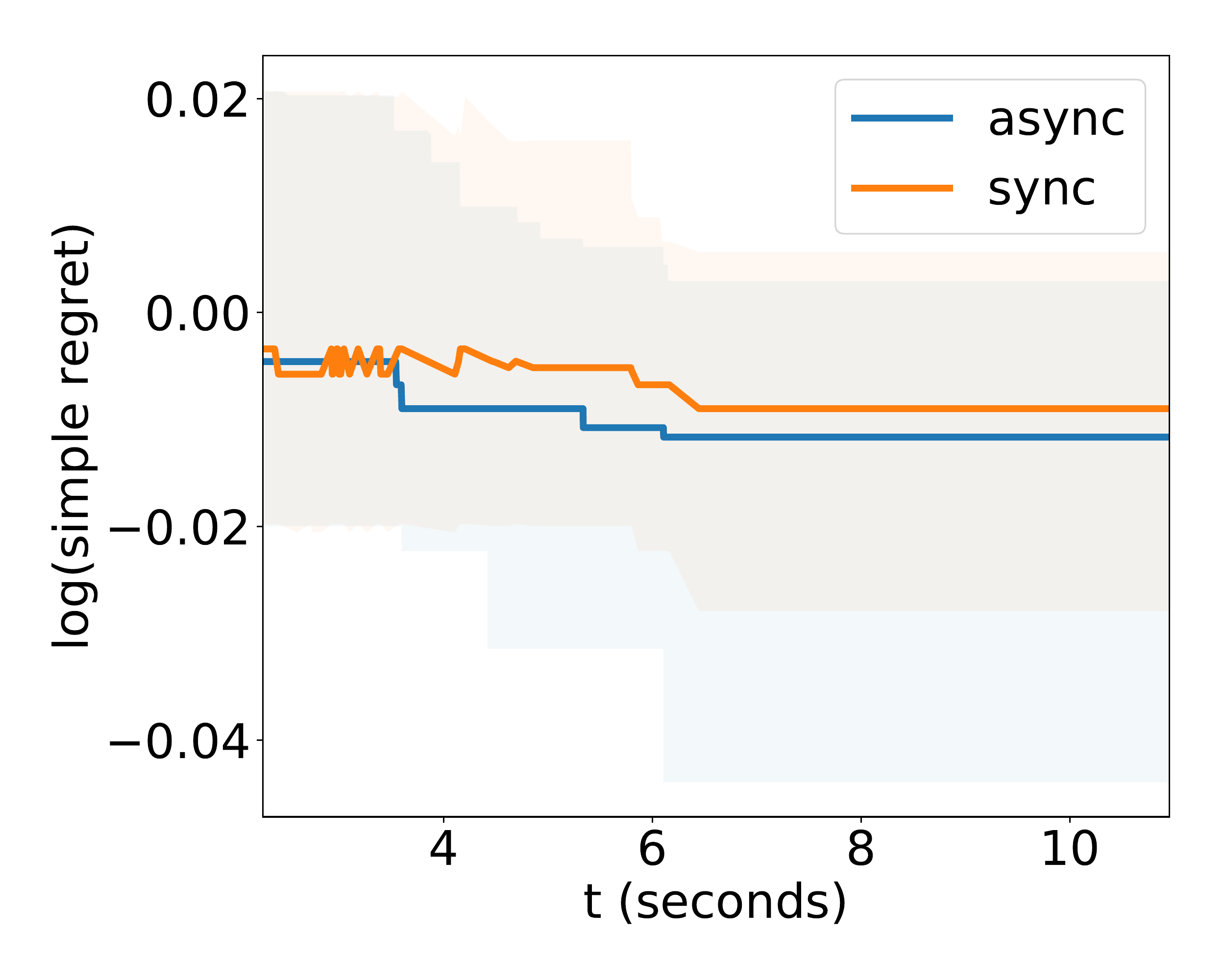}
        \caption{TS, $k = 16$}
        \label{fig:sva_time_ts_16}
    \end{subfigure}
    \begin{subfigure}[b]{0.24\textwidth}
        \includegraphics[width=\textwidth]{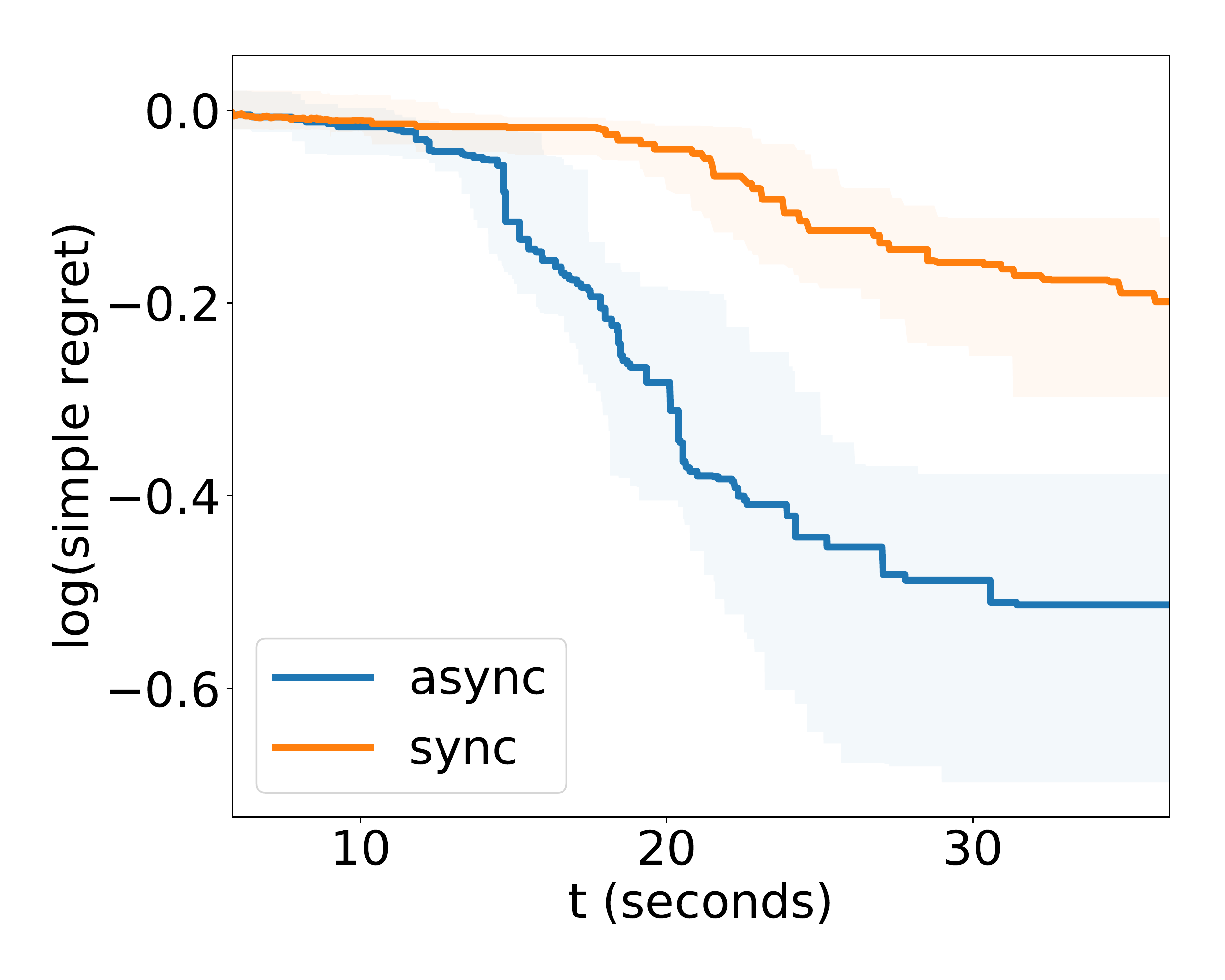}
        \caption{PLAyBOOK-HL, $k = 4$}
        \label{fig:sva_time_hllp_4}
    \end{subfigure}
    \begin{subfigure}[b]{0.24\textwidth}
        \includegraphics[width=\textwidth]{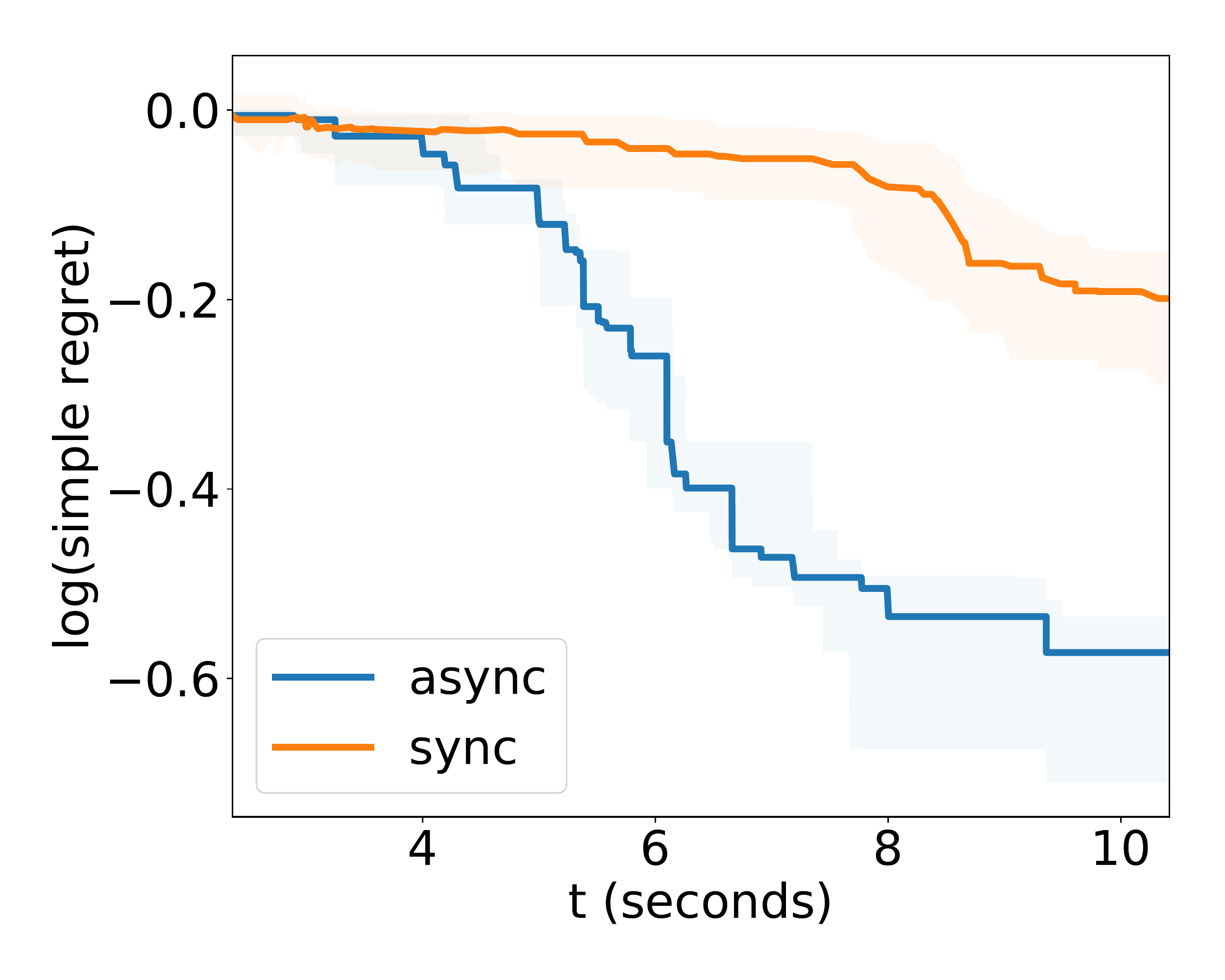}
        \caption{PLAyBOOK-HL, $k = 16$}
        \label{fig:sva_time_hllp_16}
    \end{subfigure}
    
    \begin{subfigure}[b]{0.24\textwidth}
        \includegraphics[trim={1cm 1cm 1cm 1cm}, clip, width=\textwidth]{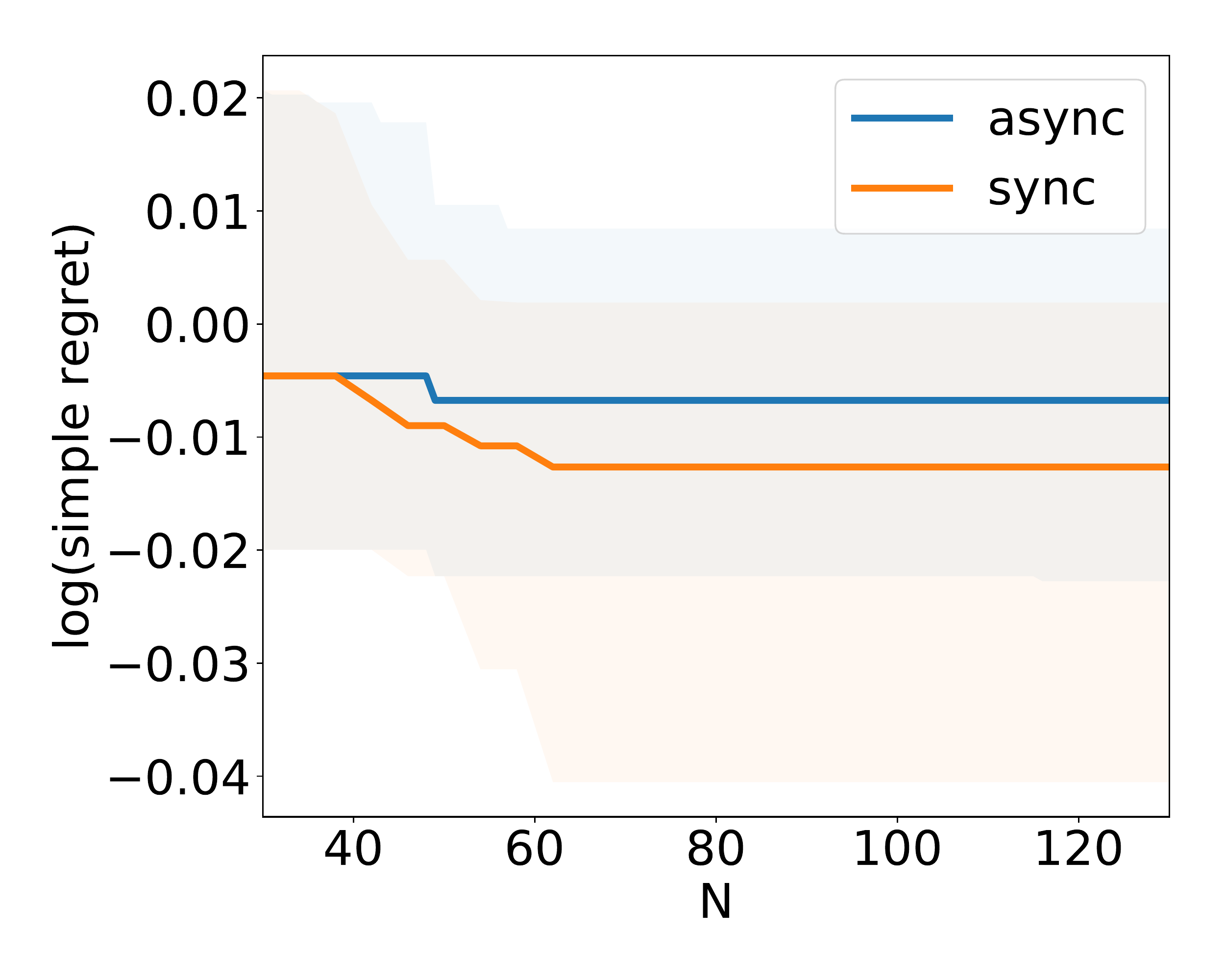}
        \caption{TS, $k = 4$}
        \label{fig:sva_ts_4}
    \end{subfigure}
    \begin{subfigure}[b]{0.24\textwidth}
        \includegraphics[width=\textwidth]{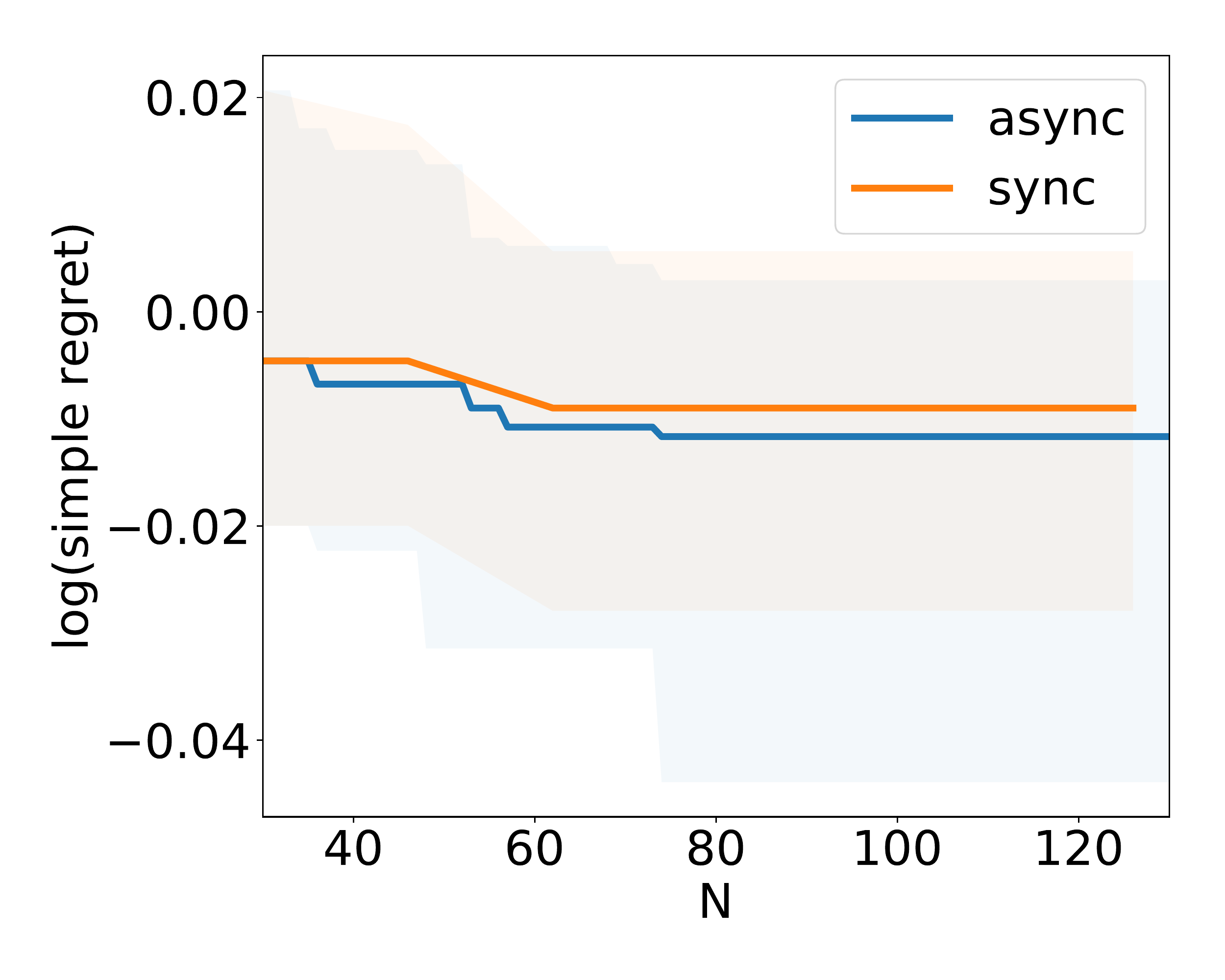}
        \caption{TS, $k = 16$}
        \label{fig:sva_ts_16}
    \end{subfigure}
    \begin{subfigure}[b]{0.24\textwidth}
        \includegraphics[width=\textwidth]{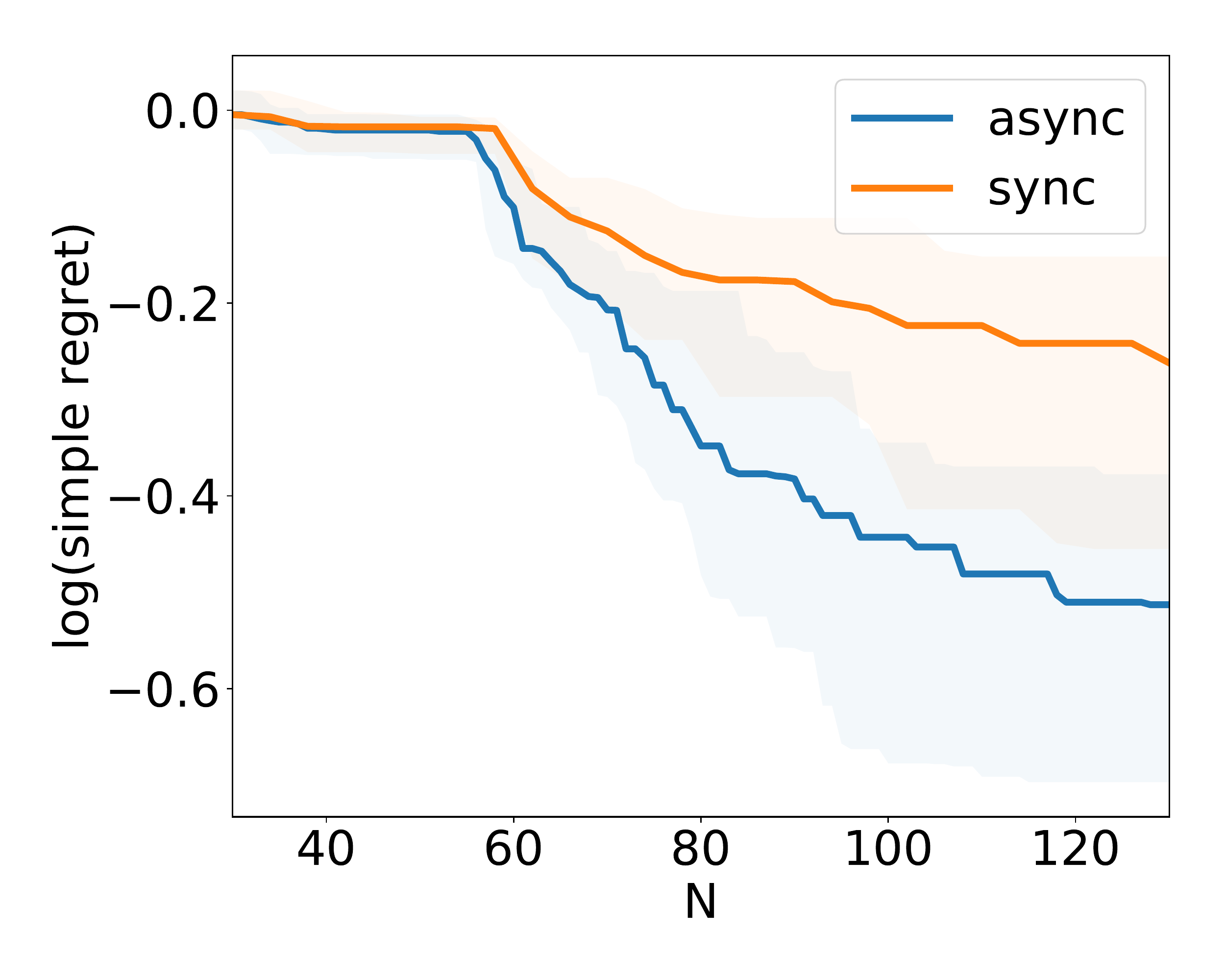}
        \caption{PLAyBOOK-HL, $k = 4$}
        \label{fig:sva_hllp_4}
    \end{subfigure}
    \begin{subfigure}[b]{0.24\textwidth}
        \includegraphics[width=\textwidth]{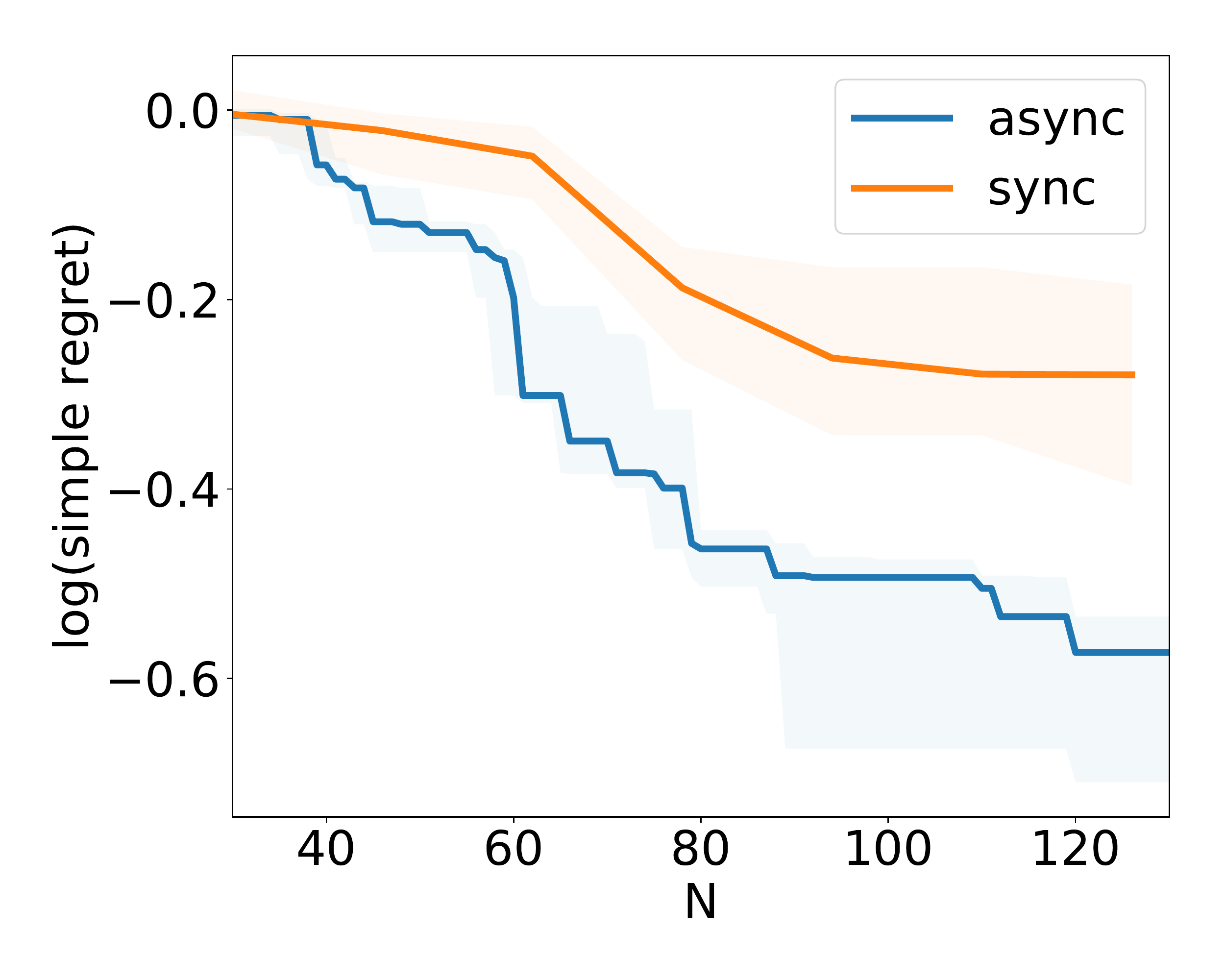}
        \caption{PLAyBOOK-HL, $k = 16$}
        \label{fig:sva_hllp_16}
    \end{subfigure}
    \caption{Head-to-head comparison on ack-10}
    \label{fig:sva_ac-10_1}
\end{figure*}

\begin{figure*}[htb!]
    \centering
        \begin{subfigure}[b]{0.24\textwidth}
        \includegraphics[trim={1cm 1cm 1cm 1cm}, clip, width=\textwidth]{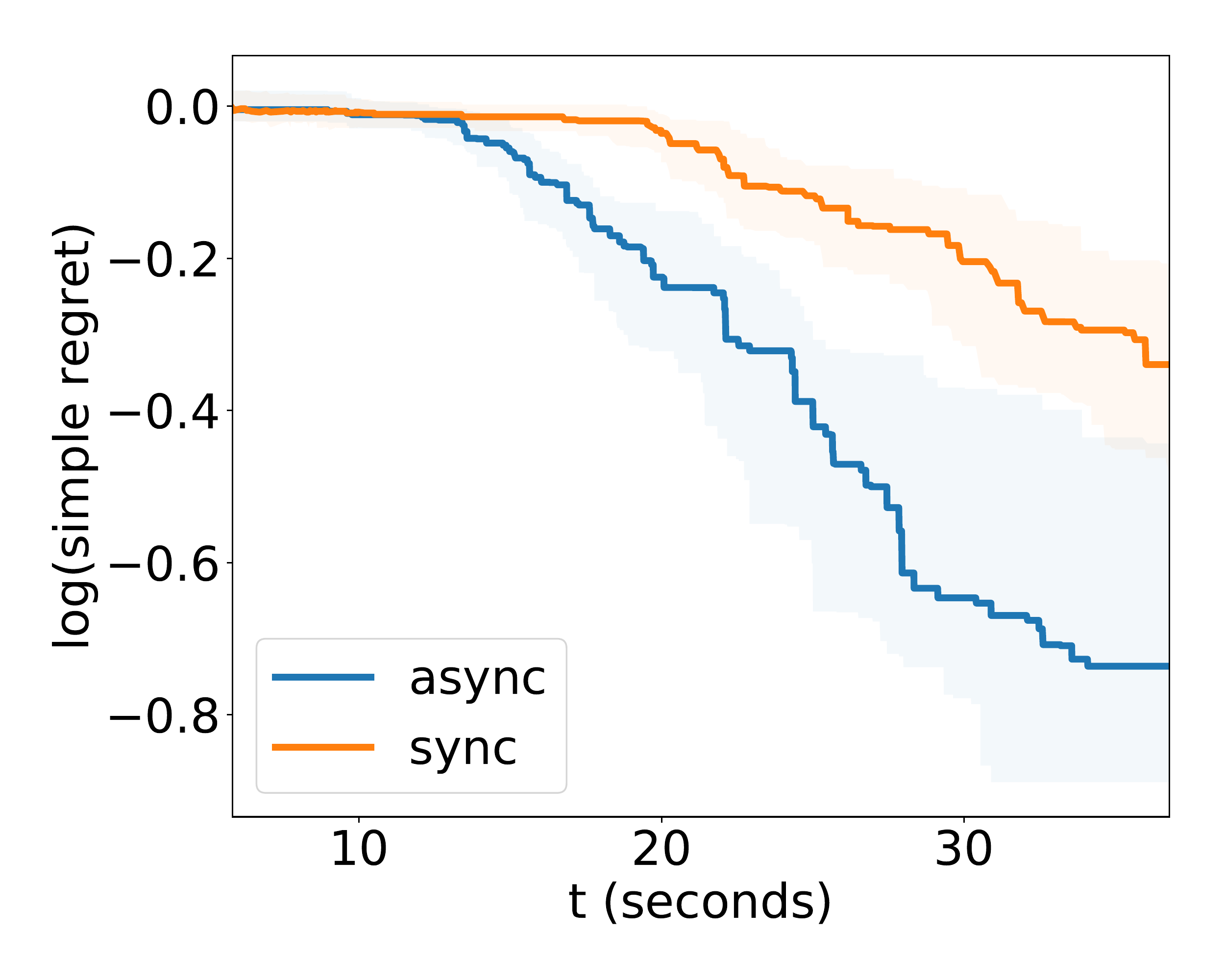}
        \caption{PLAyBOOK-L, $k = 4$}
        \label{fig:sva_time_LP_4}
    \end{subfigure}
    \begin{subfigure}[b]{0.24\textwidth}
        \includegraphics[width=\textwidth]{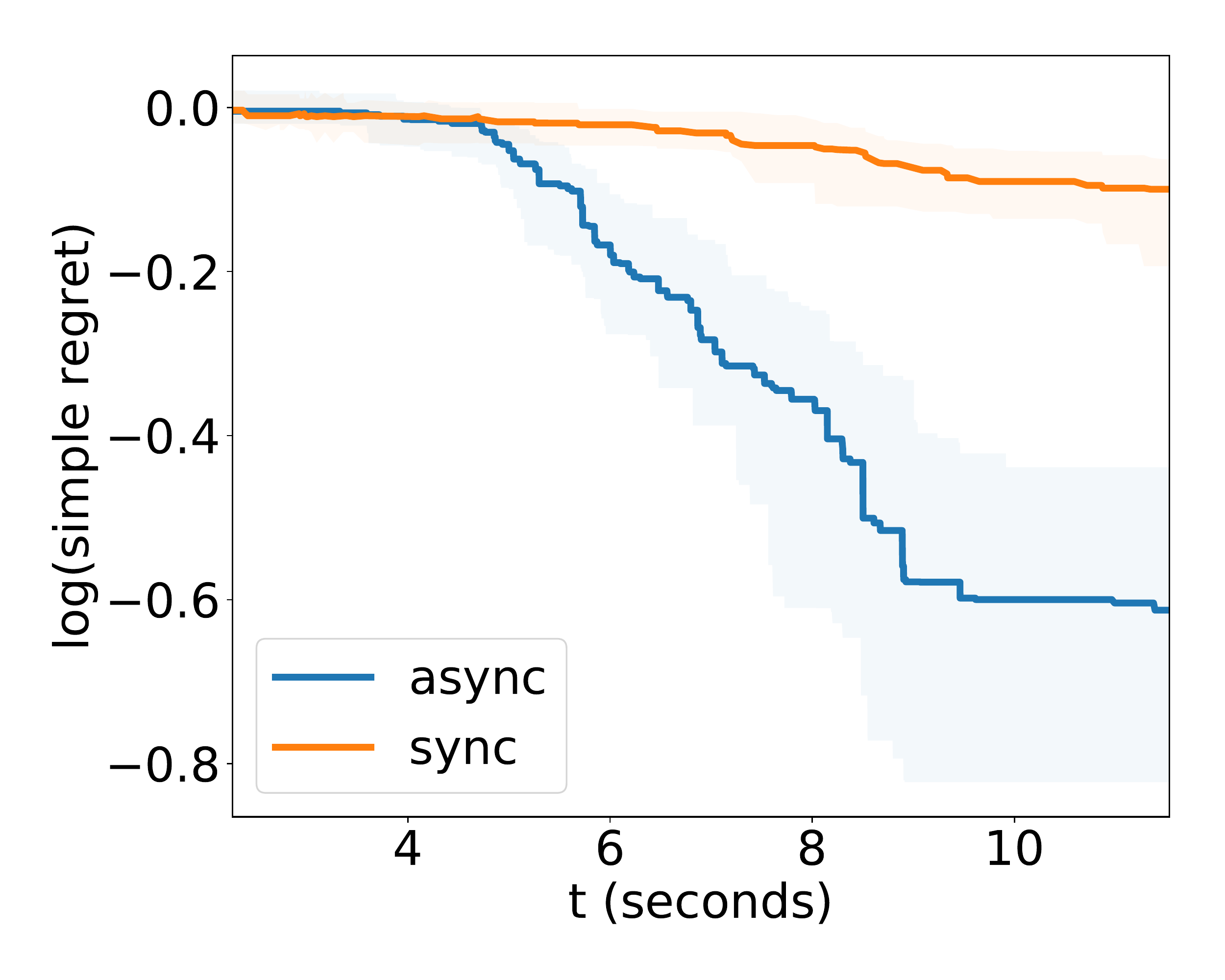}
        \caption{PLAyBOOK-L, $k = 16$}
        \label{fig:sva_time_LP_16}
    \end{subfigure}
    \begin{subfigure}[b]{0.24\textwidth}
        \includegraphics[width=\textwidth]{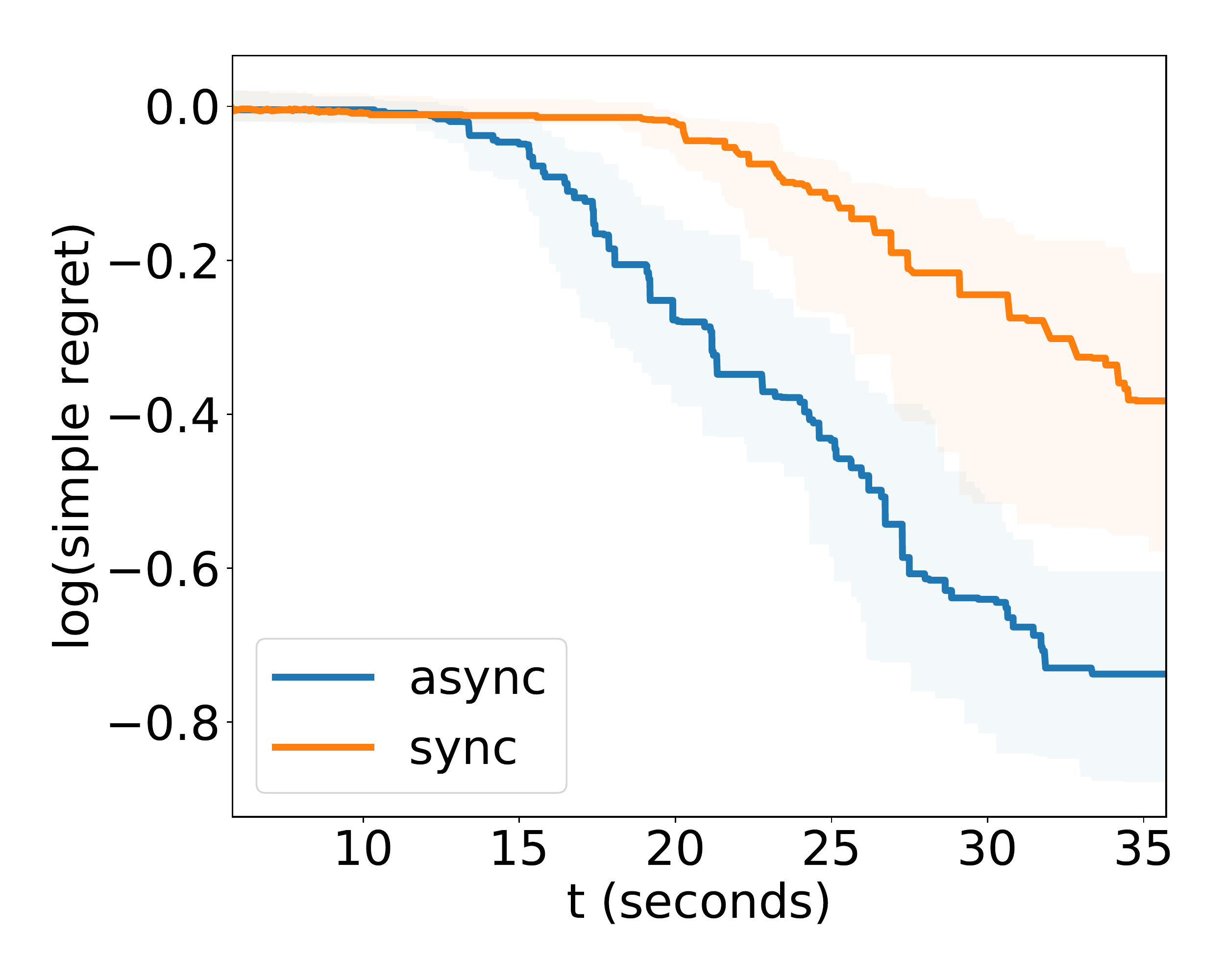}
        \caption{KB, $k = 4$}
        \label{fig:sva_time_KB_4}
    \end{subfigure}
    \begin{subfigure}[b]{0.24\textwidth}
        \includegraphics[width=\textwidth]{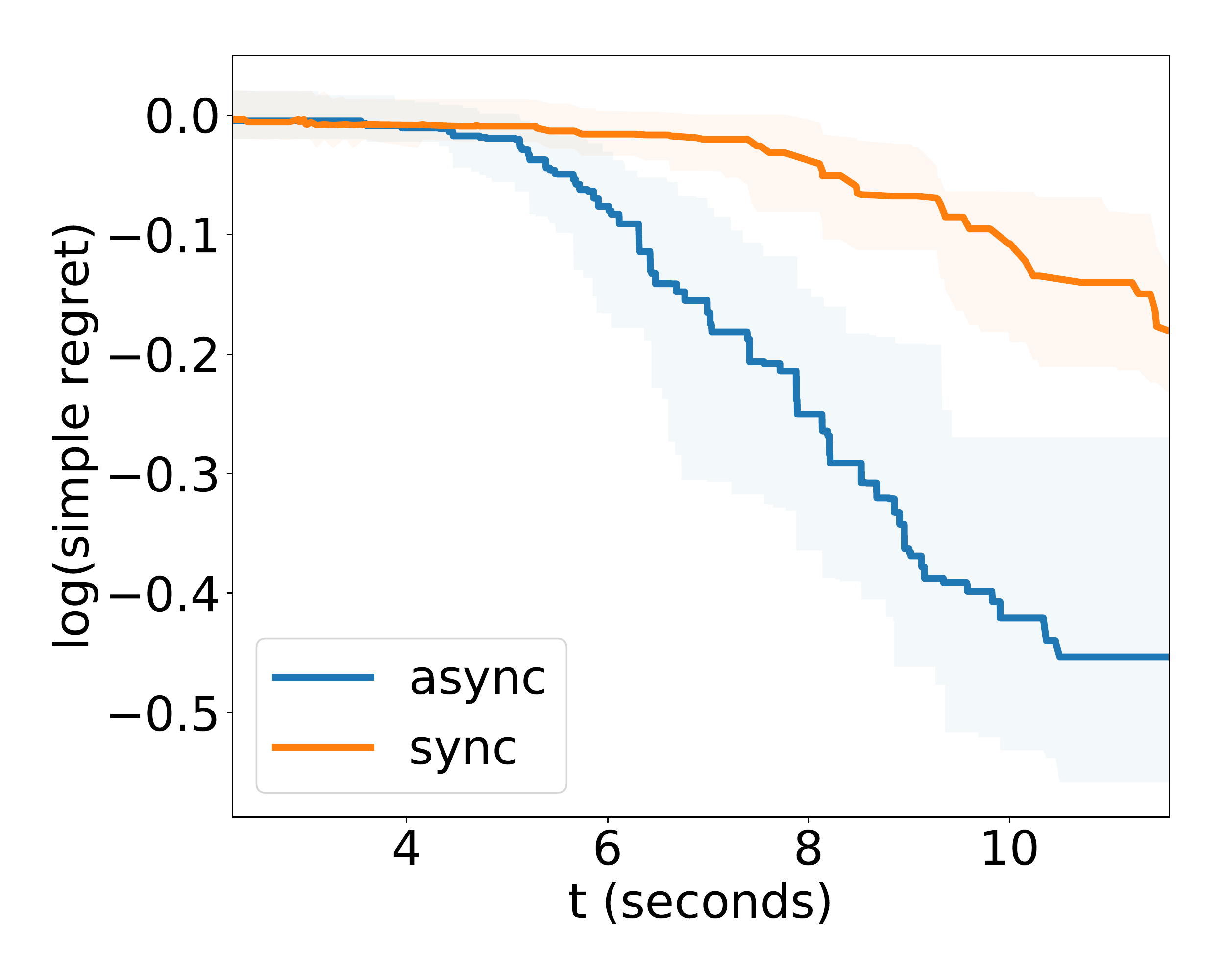}
        \caption{KB, $k = 16$}
        \label{fig:sva_time_KB_16}
    \end{subfigure}
    
    \begin{subfigure}[b]{0.24\textwidth}
        \includegraphics[trim={1cm 1cm 1cm 1cm}, clip, width=\textwidth]{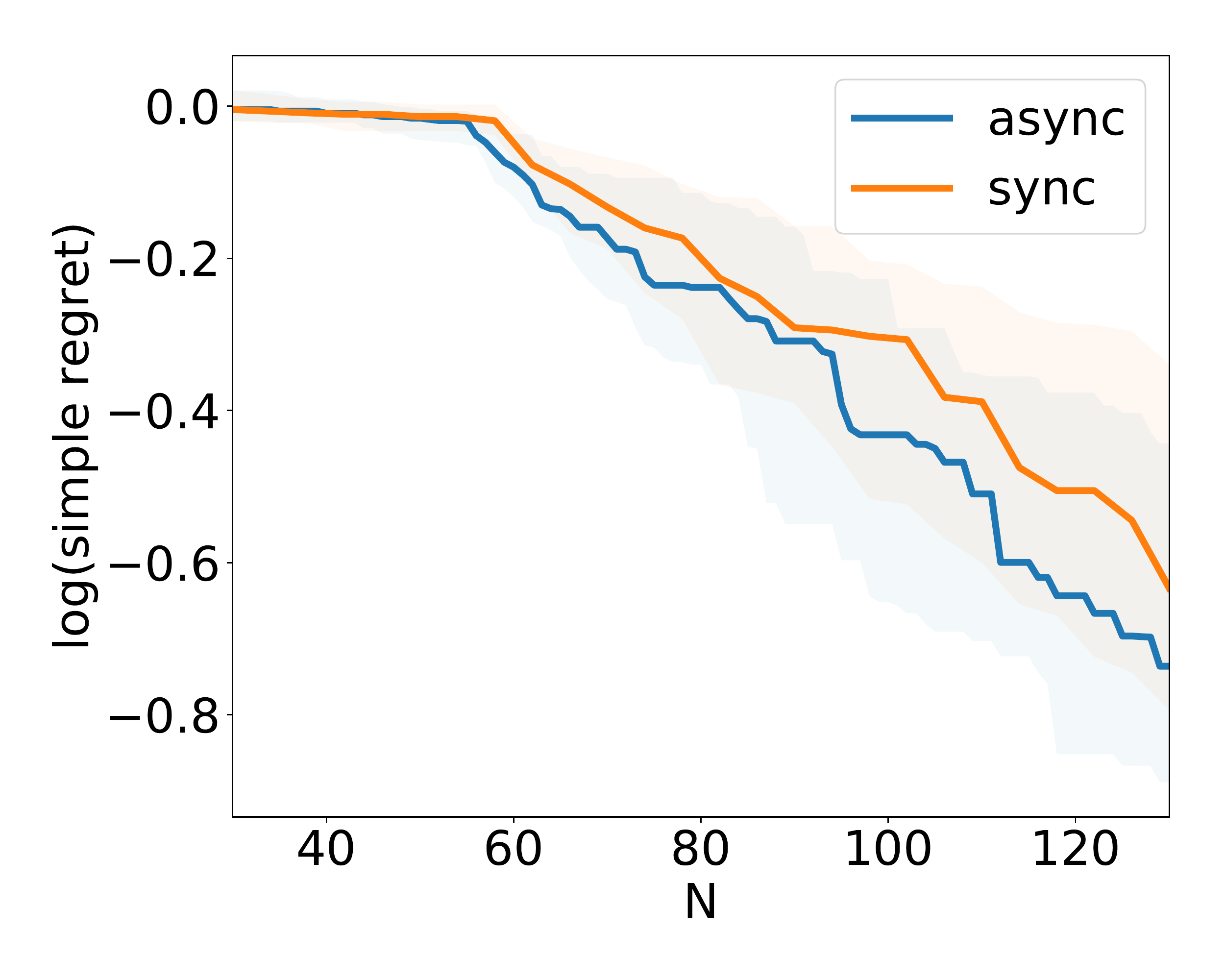}
        \caption{PLAyBOOK-L, $k = 4$}
        \label{fig:sva_LP_4}
    \end{subfigure}
    \begin{subfigure}[b]{0.24\textwidth}
        \includegraphics[width=\textwidth]{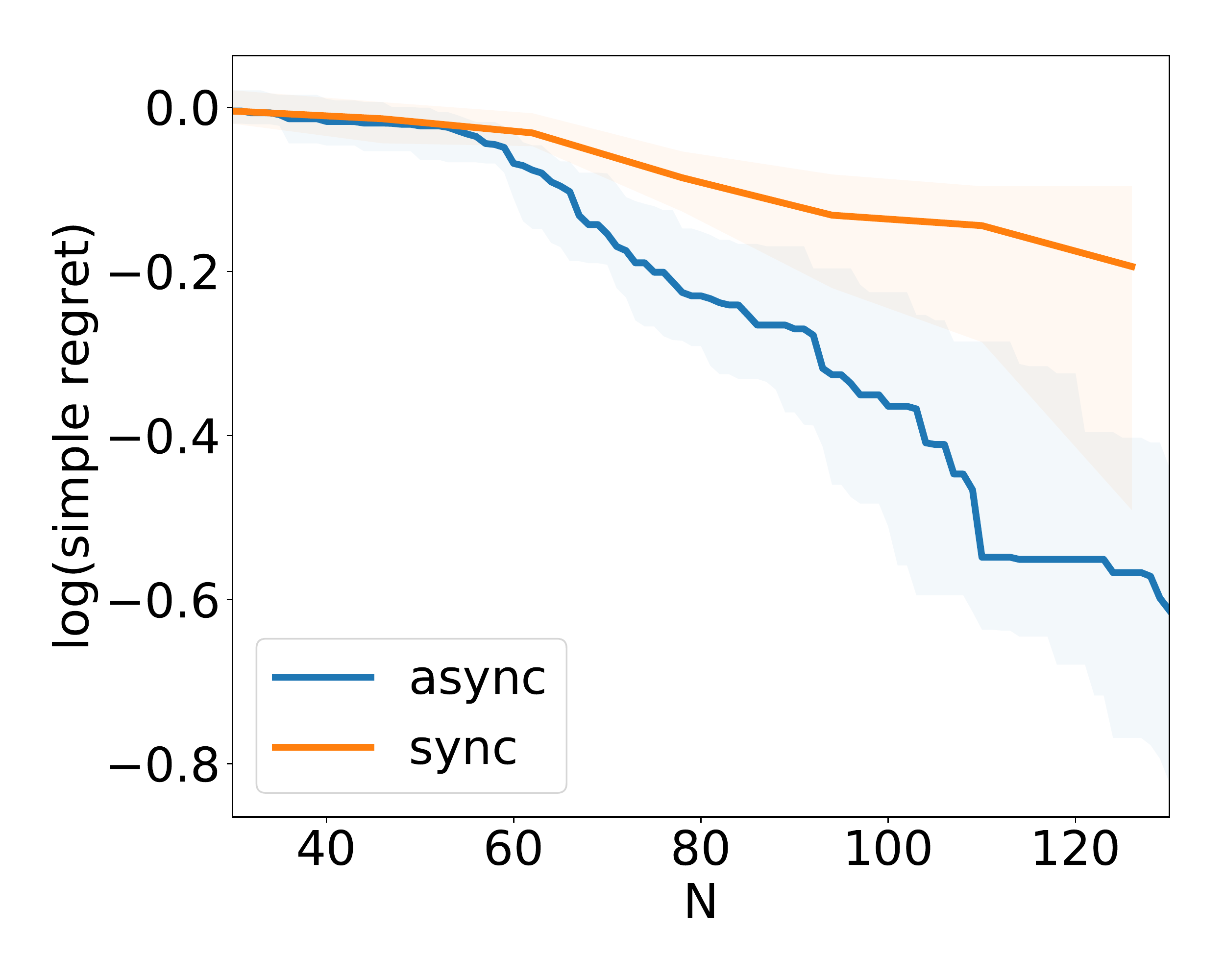}
        \caption{PLAyBOOK-L, $k = 16$}
        \label{fig:sva_LP_16}
    \end{subfigure}
    \begin{subfigure}[b]{0.24\textwidth}
        \includegraphics[width=\textwidth]{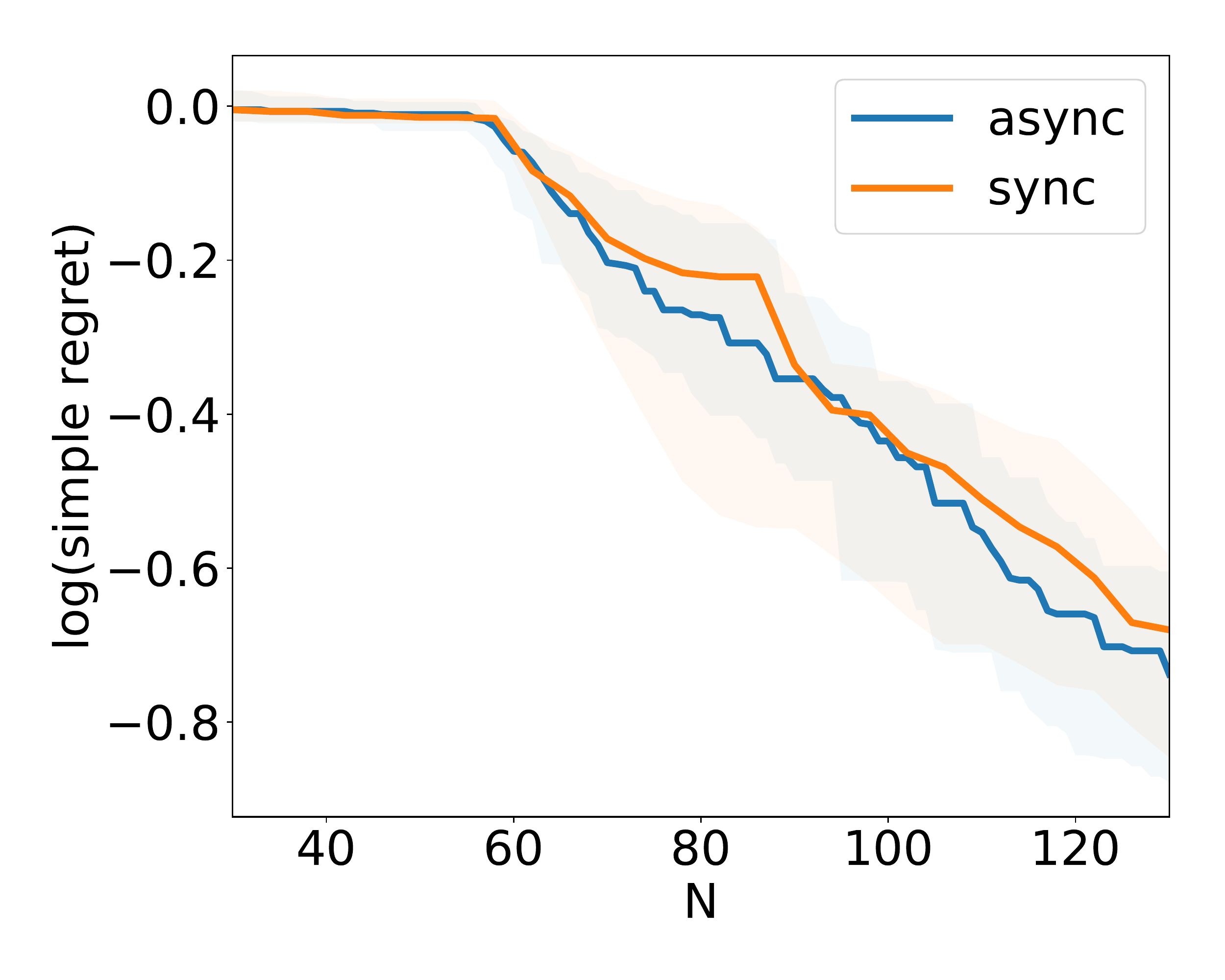}
        \caption{KB, $k = 4$}
        \label{fig:sva_KB_4}
    \end{subfigure}
    \begin{subfigure}[b]{0.24\textwidth}
        \includegraphics[width=\textwidth]{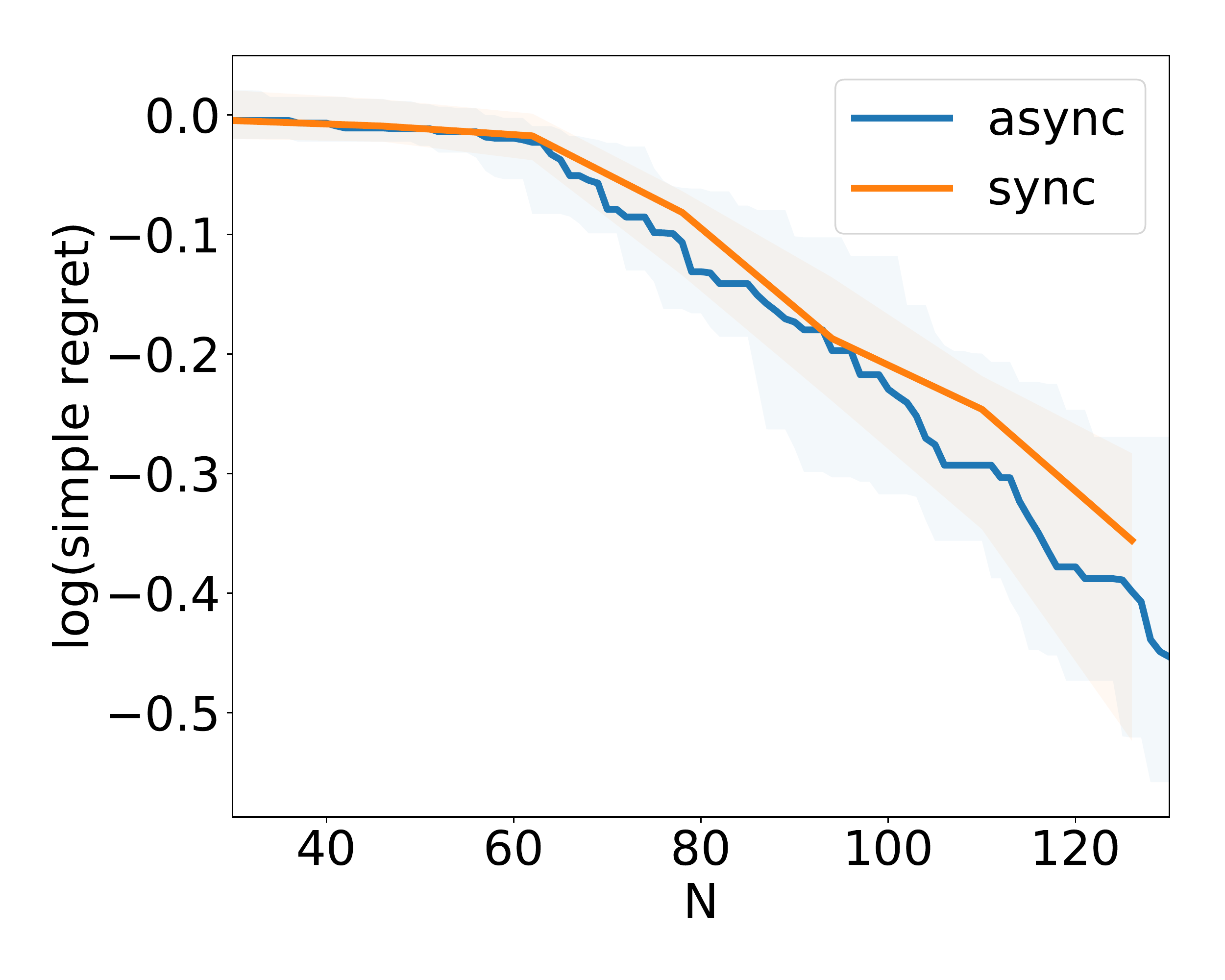}
        \caption{KB, $k = 16$}
        \label{fig:sva_KB_16}
    \end{subfigure}
    \caption{Head-to-head comparison on ack-10}
    \label{fig:sva_ac-10_2}
\end{figure*}

Similar to Fig. 4 in the main text, Figs. \ref{fig:sva_ac-10_1} and \ref{fig:sva_ac-10_2} show head-to-head comparisons of synchronous and asynchronous methods, here on the ack-10 task.

\section{Asynchronous BO}
We conducted a large number of experiments testing the different approaches for asynchronous BO. We computed the mean and standard deviation of the log of the simple regret across 30 random initialisations (see Eq. 13 in the main text for the definition). Tables \ref{tab:async_results_50}, \ref{tab:async_results_75} and \ref{tab:async_results_100} show the results after 50, 75 and 100 asynchronous BO steps respectively.

Across all of these experiments, we can see that PLAyBOOK is performing competitively, making it an attractive choice for asynchronous BO problems.

\begin{table*}[] 
\centering
\small
\begin{tabular}{llrrrrrr}
\toprule
 \multicolumn{1}{c} {\multirow{2}{*}{k}} & \multicolumn{1}{c} {\multirow{2}{*}{Task}} &\multicolumn{1}{c} { \multirow{2}{*}{KB}} & \multicolumn{1}{c} {\multirow{2}{*}{TS}} & \multicolumn{4}{c}{PLAyBOOK} \\\cmidrule{5-8} 
 &  &  &  & \multicolumn{1}{c}{L} & \multicolumn{1}{c}{LL} & \multicolumn{1}{c}{H} & \multicolumn{1}{c}{HL} \\ \midrule
2  & ack-10 &  -0.30 (0.16) &  -0.01 (0.03) &  -0.23 (0.15) & \textbf{-0.37 (0.25)} &  -0.32 (0.18) &   -0.27 (0.18) \\
   & ack-5 &  -0.55 (0.39) &  -0.22 (0.20) & \textbf{-0.85 (0.52)} &  -0.52 (0.40) &  -0.64 (0.39) &   -0.71 (0.48) \\
   & egg-2 &   0.28 (0.72) &   0.89 (0.92) &  -0.12 (1.26) &  -0.05 (1.03) & \textbf{-0.44 (2.13)} &   -0.31 (1.62) \\
   & mat-2 &   0.81 (0.30) &   1.05 (0.26) & \textbf{0.78 (0.28)} &   0.82 (0.26) &   0.81 (0.21) &    0.89 (0.20) \\
   & mat-6 &   1.02 (0.16) &   1.13 (0.14) & \textbf{0.85 (0.43)} &   0.92 (0.33) &   0.88 (0.26) &    0.86 (0.31) \\
   & mic-10 &   1.84 (0.08) &   1.92 (0.07) & \textbf{1.78 (0.11)} &   1.80 (0.09) &   1.82 (0.11) &    1.83 (0.10) \\
   & mic-5 &   0.71 (0.28) &   1.02 (0.13) &   0.67 (0.27) & \textbf{0.66 (0.26)} &   0.69 (0.33) &    0.87 (0.16) \\
   & nrobot-4 &  -0.71 (1.05) &  -0.52 (1.21) &  -0.88 (0.91) &  -0.78 (0.86) & \textbf{-0.89 (0.69)} &   -0.87 (0.91) \\
4  & ack-10 &  -0.29 (0.17) &  -0.01 (0.03) &  -0.26 (0.16) &  -0.32 (0.17) &  -0.25 (0.14) & \textbf{-0.34 (0.18)} \\ 
   & ack-5 &  -0.54 (0.36) &  -0.21 (0.13) &  -0.66 (0.45) &  -0.41 (0.27) &  -0.53 (0.41) & \textbf{-0.71 (0.52)} \\ 
   & egg-2 &   0.19 (1.06) &   0.79 (0.92) &   0.53 (0.91) &   0.23 (0.98) &   0.29 (0.86) & \textbf{-0.06 (1.26)} \\ 
   & mat-2 & \textbf{0.77 (0.30)} &   0.98 (0.37) &   0.94 (0.22) &   1.01 (0.28) &   0.82 (0.25) &    0.87 (0.22) \\
   & mat-6 &   1.06 (0.17) &   1.13 (0.05) &   0.99 (0.18) &   1.01 (0.26) &   0.95 (0.23) & \textbf{0.94 (0.20)} \\ 
   & mic-10 &   1.82 (0.11) &   1.92 (0.07) &   1.82 (0.08) &   1.83 (0.08) & \textbf{1.78 (0.09)} &    1.80 (0.08) \\
   & mic-5 & \textbf{0.66 (0.28)} &   1.01 (0.16) &   0.78 (0.23) &   0.87 (0.19) &   0.76 (0.20) &    0.82 (0.33) \\
   & nrobot-4 &  -0.67 (1.00) &  -0.39 (1.04) & \textbf{-1.01 (0.94)} &  -0.94 (0.97) &  -0.75 (0.70) &   -0.63 (0.79) \\
6  & ack-10 &  -0.24 (0.19) &  -0.01 (0.04) &  -0.29 (0.15) &  -0.25 (0.14) &  -0.20 (0.14) & \textbf{-0.31 (0.16)} \\ 
   & ack-5 &  -0.51 (0.28) &  -0.20 (0.20) & \textbf{-0.54 (0.40)} &  -0.27 (0.18) &  -0.35 (0.24) &   -0.49 (0.27) \\
   & egg-2 &   0.37 (1.05) &   0.78 (0.88) &   0.71 (0.83) &   0.77 (0.82) & \textbf{0.07 (1.25)} &    0.32 (1.11) \\
   & mat-2 &   0.85 (0.25) &   1.04 (0.19) &   0.98 (0.34) &   1.06 (0.23) & \textbf{0.84 (0.23)} &    0.85 (0.25) \\
   & mat-6 &   1.01 (0.17) &   1.07 (0.18) & \textbf{0.95 (0.28)} &   1.02 (0.28) &   1.03 (0.16) &    0.99 (0.24) \\
   & mic-10 &   1.84 (0.08) &   1.92 (0.07) &   1.84 (0.08) &   1.81 (0.08) &   1.84 (0.08) & \textbf{1.80 (0.09)} \\ 
   & mic-5 & \textbf{0.76 (0.21)} &   1.04 (0.15) &   0.84 (0.23) &   0.89 (0.24) &   0.88 (0.19) &    0.87 (0.20) \\
   & nrobot-4 &  -0.58 (0.96) &  -0.22 (0.95) & \textbf{-0.69 (1.04)} &  -0.47 (0.85) &  -0.63 (1.02) &   -0.51 (0.99) \\
8  & ack-10 &  -0.23 (0.17) &  -0.01 (0.03) &  -0.26 (0.17) &  -0.34 (0.15) &  -0.17 (0.11) & \textbf{-0.40 (0.14)} \\ 
   & ack-5 &  -0.44 (0.27) &  -0.19 (0.14) &  -0.65 (0.40) &  -0.33 (0.21) &  -0.38 (0.30) & \textbf{-0.68 (0.37)} \\ 
   & egg-2 &   0.31 (0.93) &   0.65 (0.92) &   0.60 (1.20) &   0.93 (0.77) &   0.42 (0.61) & \textbf{0.17 (1.05)} \\ 
   & mat-2 &   0.80 (0.22) &   0.93 (0.26) &   0.81 (0.44) &   1.09 (0.20) &   0.87 (0.27) & \textbf{0.79 (0.30)} \\ 
   & mat-6 & \textbf{1.01 (0.19)} &   1.13 (0.06) &   1.03 (0.17) &   1.02 (0.22) &   1.03 (0.15) &    1.03 (0.20) \\
   & mic-10 &   1.84 (0.09) &   1.91 (0.07) &   1.82 (0.08) &   1.81 (0.10) &   1.80 (0.13) & \textbf{1.78 (0.08)} \\ 
   & mic-5 &   0.70 (0.33) &   1.02 (0.20) & \textbf{0.68 (0.32)} &   0.79 (0.25) &   0.73 (0.32) &    0.69 (0.32) \\
   & nrobot-4 &  -0.61 (0.78) &  -0.20 (0.96) & \textbf{-0.94 (0.92)} &  -0.62 (1.14) &  -0.76 (1.11) &   -0.39 (0.73) \\
16 & ack-10 &  -0.13 (0.08) &  -0.02 (0.04) &  -0.24 (0.13) &  -0.24 (0.13) &  -0.22 (0.10) & \textbf{-0.44 (0.13)} \\ 
   & ack-5 &  -0.36 (0.30) &  -0.18 (0.16) &  -0.49 (0.34) &  -0.36 (0.24) &  -0.28 (0.19) & \textbf{-0.73 (0.26)} \\ 
   & egg-2 &   0.58 (0.73) &   0.56 (1.10) &   1.04 (0.51) &   1.22 (0.50) & \textbf{0.30 (0.96)} &    0.71 (0.53) \\
   & mat-2 & \textbf{0.82 (0.27)} &   0.89 (0.24) &   1.07 (0.17) &   1.12 (0.20) &   0.92 (0.31) &    0.90 (0.31) \\
   & mat-6 &   1.10 (0.05) &   1.14 (0.03) &   1.05 (0.14) &   1.06 (0.21) &   1.02 (0.26) & \textbf{1.00 (0.26)} \\ 
   & mic-10 &   1.86 (0.08) &   1.90 (0.07) &   1.82 (0.07) &   1.82 (0.09) &   1.83 (0.08) & \textbf{1.79 (0.09)} \\ 
   & mic-5 & \textbf{0.81 (0.25)} &   1.02 (0.19) &   0.85 (0.21) &   0.84 (0.23) &   0.81 (0.22) &    0.82 (0.25) \\
   & nrobot-4 &  -0.43 (0.94) &  -0.22 (0.99) & \textbf{-0.78 (0.95)} &  -0.14 (0.77) &  -0.37 (0.80) &   -0.40 (0.81) \\
\bottomrule
\end{tabular}
\caption{Mean and standard deviation of the log(regret) after 50 steps of asynchronous BO.}
\label{tab:async_results_50}
\end{table*} 

\begin{table*}[] 
\centering
\small
\begin{tabular}{llrrrrrr}
\toprule
 \multicolumn{1}{c} {\multirow{2}{*}{k}} & \multicolumn{1}{c} {\multirow{2}{*}{Task}} &\multicolumn{1}{c} { \multirow{2}{*}{KB}} & \multicolumn{1}{c} {\multirow{2}{*}{TS}} & \multicolumn{4}{c}{PLAyBOOK} \\\cmidrule{5-8} 
 &  &  &  & \multicolumn{1}{c}{L} & \multicolumn{1}{c}{LL} & \multicolumn{1}{c}{H} & \multicolumn{1}{c}{HL} \\ 
 \midrule
2  & ack-10 &  -0.43 (0.18) &  -0.01 (0.03) &  -0.48 (0.27) & \textbf{-0.58 (0.26)} &  -0.55 (0.32) &   -0.45 (0.28) \\
   & ack-5 &  -0.91 (0.56) &  -0.32 (0.22) & \textbf{-1.15 (0.58)} &  -0.76 (0.50) &  -1.03 (0.52) &   -0.92 (0.51) \\
   & egg-2 &  -0.12 (0.92) &   0.87 (0.91) &  -0.59 (1.16) & \textbf{-1.13 (2.14)} &  -0.81 (1.99) &   -0.82 (1.68) \\
   & mat-2 &   0.80 (0.30) &   1.05 (0.27) & \textbf{0.76 (0.28)} &   0.81 (0.26) &   0.81 (0.21) &    0.87 (0.20) \\
   & mat-6 &   0.95 (0.21) &   1.17 (0.14) & \textbf{0.74 (0.53)} &   0.87 (0.31) &   0.84 (0.34) &    0.89 (0.29) \\
   & mic-10 &   1.79 (0.12) &   1.92 (0.07) & \textbf{1.75 (0.11)} &   1.76 (0.14) &   1.79 (0.12) &    1.79 (0.13) \\
   & mic-5 & \textbf{0.52 (0.42)} &   0.97 (0.17) &   0.61 (0.29) &   0.56 (0.24) &   0.57 (0.37) &    0.73 (0.28) \\
   & nrobot-4 &  -1.06 (1.08) &  -0.83 (1.18) &  -1.20 (0.86) & \textbf{-1.31 (0.75)} &  -1.24 (0.78) &   -1.29 (0.80) \\
4  & ack-10 & \textbf{-0.52 (0.21)} &  -0.01 (0.03) &  -0.51 (0.27) &  -0.42 (0.19) &  -0.46 (0.22) &   -0.50 (0.22) \\
   & ack-5 &  -0.83 (0.48) &  -0.31 (0.23) & \textbf{-1.10 (0.53)} &  -0.57 (0.33) &  -0.75 (0.62) &   -0.90 (0.54) \\
   & egg-2 &  -0.19 (0.87) &   0.69 (1.13) &   0.16 (1.70) &  -0.81 (2.55) &  -0.23 (1.10) & \textbf{-0.89 (2.48)} \\ 
   & mat-2 & \textbf{0.75 (0.29)} &   0.98 (0.37) &   0.93 (0.22) &   1.01 (0.28) &   0.80 (0.25) &    0.85 (0.22) \\
   & mat-6 &   0.97 (0.27) &   1.18 (0.06) &   0.95 (0.23) &   1.02 (0.24) & \textbf{0.86 (0.28)} &    0.87 (0.30) \\
   & mic-10 &   1.77 (0.12) &   1.92 (0.07) &   1.77 (0.09) &   1.78 (0.09) & \textbf{1.74 (0.09)} &    1.77 (0.10) \\
   & mic-5 & \textbf{0.56 (0.30)} &   0.97 (0.14) &   0.69 (0.25) &   0.76 (0.19) &   0.63 (0.22) &    0.67 (0.34) \\
   & nrobot-4 &  -0.92 (0.95) &  -1.01 (1.31) & \textbf{-1.51 (0.88)} &  -1.23 (0.89) &  -1.07 (0.68) &   -1.04 (0.79) \\
6  & ack-10 &  -0.41 (0.22) &  -0.01 (0.04) & \textbf{-0.50 (0.26)} &  -0.30 (0.15) &  -0.32 (0.18) &   -0.38 (0.19) \\
   & ack-5 &  -0.83 (0.46) &  -0.32 (0.23) & \textbf{-0.86 (0.54)} &  -0.36 (0.23) &  -0.52 (0.31) &   -0.76 (0.37) \\
   & egg-2 &   0.19 (1.00) &   0.64 (1.14) &   0.56 (0.93) &   0.75 (0.84) & \textbf{-0.34 (1.48)} &   -0.21 (1.03) \\
   & mat-2 & \textbf{0.82 (0.26)} &   1.03 (0.20) &   0.98 (0.34) &   1.06 (0.23) &   0.82 (0.22) &    0.84 (0.25) \\
   & mat-6 &   0.92 (0.25) &   1.14 (0.17) & \textbf{0.86 (0.36)} &   1.08 (0.25) &   0.95 (0.28) &    0.94 (0.26) \\
   & mic-10 &   1.81 (0.08) &   1.92 (0.07) &   1.81 (0.08) &   1.79 (0.09) &   1.80 (0.10) & \textbf{1.78 (0.10)} \\ 
   & mic-5 & \textbf{0.63 (0.28)} &   1.00 (0.14) &   0.74 (0.29) &   0.82 (0.22) &   0.76 (0.26) &    0.69 (0.33) \\
   & nrobot-4 &  -1.02 (0.88) &  -0.73 (1.20) &  -0.99 (1.04) &  -0.78 (0.96) & \textbf{-1.04 (1.02)} &   -1.02 (0.95) \\
8  & ack-10 &  -0.44 (0.21) &  -0.01 (0.03) &  -0.47 (0.21) &  -0.40 (0.20) &  -0.28 (0.17) & \textbf{-0.52 (0.16)} \\ 
   & ack-5 &  -0.77 (0.40) &  -0.33 (0.20) & \textbf{-1.04 (0.45)} &  -0.39 (0.22) &  -0.52 (0.41) &   -0.92 (0.36) \\
   & egg-2 & \textbf{-0.11 (0.96)} &   0.57 (1.00) &   0.40 (1.22) &   0.85 (0.84) &   0.22 (0.51) &   -0.05 (0.96) \\
   & mat-2 &   0.78 (0.23) &   0.91 (0.26) &   0.81 (0.44) &   1.09 (0.20) &   0.82 (0.26) & \textbf{0.76 (0.30)} \\ 
   & mat-6 &   0.95 (0.24) &   1.18 (0.10) &   0.97 (0.20) &   1.10 (0.20) & \textbf{0.93 (0.25)} &    0.97 (0.28) \\
   & mic-10 &   1.80 (0.09) &   1.91 (0.07) & \textbf{1.75 (0.09)} &   1.78 (0.09) &   1.78 (0.13) &    1.76 (0.08) \\
   & mic-5 &   0.58 (0.38) &   0.98 (0.19) &   0.59 (0.36) &   0.70 (0.26) &   0.61 (0.33) & \textbf{0.55 (0.53)} \\ 
   & nrobot-4 &  -0.92 (0.89) &  -0.65 (1.01) & \textbf{-1.25 (0.82)} &  -0.92 (1.08) &  -1.07 (1.10) &   -0.92 (0.88) \\
16 & ack-10 &  -0.27 (0.14) &  -0.02 (0.04) &  -0.43 (0.21) &  -0.31 (0.20) &  -0.31 (0.14) & \textbf{-0.54 (0.16)} \\ 
   & ack-5 &  -0.55 (0.34) &  -0.29 (0.19) &  -0.77 (0.42) &  -0.39 (0.25) &  -0.39 (0.28) & \textbf{-0.94 (0.30)} \\ 
   & egg-2 &   0.26 (0.74) &   0.41 (1.18) &   0.89 (0.63) &   1.21 (0.50) & \textbf{-0.39 (1.81)} &    0.42 (0.61) \\
   & mat-2 & \textbf{0.79 (0.27)} &   0.88 (0.24) &   1.07 (0.17) &   1.12 (0.20) &   0.87 (0.30) &    0.86 (0.31) \\
   & mat-6 &   1.03 (0.20) &   1.20 (0.03) &   0.95 (0.21) &   1.08 (0.25) & \textbf{0.94 (0.26)} &    0.96 (0.36) \\
   & mic-10 &   1.83 (0.09) &   1.90 (0.07) &   1.79 (0.08) &   1.79 (0.09) &   1.79 (0.10) & \textbf{1.75 (0.13)} \\ 
   & mic-5 & \textbf{0.66 (0.27)} &   0.95 (0.21) &   0.77 (0.25) &   0.73 (0.39) &   0.75 (0.23) &    0.75 (0.25) \\
   & nrobot-4 &  -0.72 (0.87) &  -0.80 (1.24) & \textbf{-1.00 (1.01)} &  -0.61 (0.95) &  -0.72 (0.61) &   -0.79 (0.78) \\
\bottomrule
\end{tabular}
\caption{Mean and standard deviation of the log(regret) after 75 steps of asynchronous BO.}
\label{tab:async_results_75}
\end{table*}

\begin{table*}[] 
\centering
\small
\begin{tabular}{llrrrrrr}
\toprule
 \multicolumn{1}{c} {\multirow{2}{*}{k}} & \multicolumn{1}{c} {\multirow{2}{*}{Task}} &\multicolumn{1}{c} { \multirow{2}{*}{KB}} & \multicolumn{1}{c} {\multirow{2}{*}{TS}} & \multicolumn{4}{c}{PLAyBOOK} \\\cmidrule{5-8} 
 &  &  &  & \multicolumn{1}{c}{L} & \multicolumn{1}{c}{LL} & \multicolumn{1}{c}{H} & \multicolumn{1}{c}{HL} \\ 
 \midrule
2  & ack-10 &  -0.67 (0.27) &  -0.01 (0.03) &  -0.72 (0.34) &  -0.73 (0.29) & \textbf{-0.80 (0.35)} &   -0.58 (0.26) \\
   & ack-5 &  -1.28 (0.70) &  -0.35 (0.22) & \textbf{-1.49 (0.65)} &  -0.95 (0.53) &  -1.39 (0.62) &   -1.08 (0.47) \\
   & egg-2 &  -0.42 (1.62) &   0.80 (0.92) &  -1.10 (1.56) & \textbf{-1.58 (2.49)} &  -1.54 (2.28) &   -1.30 (2.01) \\
   & mat-2 &   0.79 (0.30) &   1.04 (0.25) & \textbf{0.76 (0.28)} &   0.81 (0.26) &   0.81 (0.21) &    0.87 (0.19) \\
   & mat-6 &   1.00 (0.21) &   1.27 (0.17) & \textbf{0.93 (0.36)} &   1.00 (0.31) &   0.94 (0.26) &    1.01 (0.27) \\
   & mic-10 &   1.76 (0.13) &   1.92 (0.07) & \textbf{1.72 (0.10)} &   1.73 (0.13) &   1.72 (0.13) &    1.76 (0.12) \\
   & mic-5 & \textbf{0.39 (0.51)} &   0.94 (0.18) &   0.54 (0.31) &   0.45 (0.24) &   0.45 (0.36) &    0.65 (0.28) \\
   & nrobot-4 &  -1.33 (1.02) &  -1.12 (1.21) &  -1.54 (0.84) &  -1.51 (0.85) &  -1.46 (0.74) & \textbf{-1.58 (0.78)} \\ 
4  & ack-10 & \textbf{-0.72 (0.24)} &  -0.01 (0.03) &  -0.68 (0.28) &  -0.54 (0.24) &  -0.67 (0.24) &   -0.56 (0.24) \\
   & ack-5 &  -1.13 (0.59) &  -0.41 (0.25) & \textbf{-1.46 (0.59)} &  -0.68 (0.32) &  -0.98 (0.73) &   -1.04 (0.52) \\
   & egg-2 &  -0.27 (0.82) &   0.65 (1.12) &  -0.16 (1.71) &  -1.09 (2.57) &  -0.50 (1.03) & \textbf{-1.17 (2.53)} \\ 
   & mat-2 & \textbf{0.74 (0.29)} &   0.98 (0.37) &   0.93 (0.22) &   1.01 (0.28) &   0.79 (0.25) &    0.85 (0.22) \\
   & mat-6 &   1.03 (0.25) &   1.30 (0.06) &   1.07 (0.19) &   1.12 (0.20) & \textbf{0.98 (0.25)} &    1.01 (0.23) \\
   & mic-10 &   1.74 (0.11) &   1.92 (0.07) &   1.75 (0.09) &   1.74 (0.10) & \textbf{1.70 (0.09)} &    1.74 (0.11) \\
   & mic-5 & \textbf{0.44 (0.32)} &   0.96 (0.14) &   0.55 (0.27) &   0.63 (0.21) &   0.50 (0.27) &    0.63 (0.34) \\
   & nrobot-4 &  -1.14 (0.88) &  -1.24 (1.23) & \textbf{-1.73 (0.88)} &  -1.48 (0.72) &  -1.32 (0.83) &   -1.24 (0.83) \\
6  & ack-10 &  -0.63 (0.27) &  -0.01 (0.04) & \textbf{-0.65 (0.29)} &  -0.37 (0.20) &  -0.44 (0.25) &   -0.43 (0.20) \\
   & ack-5 & \textbf{-1.19 (0.60)} &  -0.39 (0.26) &  -1.17 (0.65) &  -0.39 (0.22) &  -0.68 (0.41) &   -0.87 (0.40) \\
   & egg-2 &  -0.18 (1.05) &   0.57 (1.15) &   0.50 (0.94) &   0.65 (0.98) & \textbf{-0.53 (1.40)} &   -0.45 (0.99) \\
   & mat-2 &   0.82 (0.25) &   1.02 (0.20) &   0.98 (0.34) &   1.06 (0.23) & \textbf{0.81 (0.22)} &    0.83 (0.24) \\
   & mat-6 &   1.02 (0.24) &   1.25 (0.16) & \textbf{0.99 (0.28)} &   1.18 (0.27) &   1.05 (0.23) &    1.03 (0.23) \\
   & mic-10 &   1.77 (0.10) &   1.92 (0.07) & \textbf{1.76 (0.08)} &   1.76 (0.08) &   1.76 (0.13) &    1.77 (0.10) \\
   & mic-5 & \textbf{0.51 (0.34)} &   0.96 (0.15) &   0.66 (0.27) &   0.72 (0.25) &   0.66 (0.37) &    0.63 (0.30) \\
   & nrobot-4 &  -1.27 (0.80) &  -1.12 (1.28) &  -1.24 (0.89) &  -1.03 (0.94) & \textbf{-1.35 (0.95)} &   -1.27 (1.00) \\
8  & ack-10 &  -0.57 (0.22) &  -0.01 (0.03) &  -0.62 (0.26) &  -0.44 (0.23) &  -0.41 (0.25) & \textbf{-0.63 (0.16)} \\ 
   & ack-5 &  -1.16 (0.47) &  -0.42 (0.24) & \textbf{-1.32 (0.46)} &  -0.47 (0.23) &  -0.70 (0.51) &   -1.12 (0.41) \\
   & egg-2 &  -0.24 (0.91) &   0.42 (1.13) &  -0.02 (2.20) &   0.78 (0.86) &  -0.04 (0.54) & \textbf{-0.27 (0.85)} \\ 
   & mat-2 &   0.78 (0.23) &   0.90 (0.25) &   0.81 (0.44) &   1.09 (0.20) &   0.81 (0.26) & \textbf{0.75 (0.30)} \\ 
   & mat-6 & \textbf{1.01 (0.23)} &   1.30 (0.09) &   1.07 (0.17) &   1.23 (0.17) &   1.02 (0.19) &    1.04 (0.25) \\
   & mic-10 &   1.78 (0.08) &   1.91 (0.07) & \textbf{1.72 (0.10)} &   1.75 (0.09) &   1.74 (0.13) &    1.74 (0.08) \\
   & mic-5 & \textbf{0.46 (0.37)} &   0.93 (0.19) &   0.53 (0.35) &   0.68 (0.26) &   0.50 (0.32) &    0.50 (0.52) \\
   & nrobot-4 &  -1.24 (0.76) &  -1.18 (1.23) & \textbf{-1.44 (0.84)} &  -1.17 (0.98) &  -1.35 (1.02) &   -1.13 (0.85) \\
16 & ack-10 &  -0.44 (0.20) &  -0.02 (0.04) & \textbf{-0.63 (0.24)} &  -0.37 (0.18) &  -0.40 (0.18) &   -0.61 (0.16) \\
   & ack-5 &  -0.83 (0.43) &  -0.41 (0.23) &  -1.03 (0.51) &  -0.41 (0.24) &  -0.43 (0.30) & \textbf{-1.11 (0.35)} \\ 
   & egg-2 &  -0.41 (2.18) &   0.34 (1.17) &   0.83 (0.67) &   1.21 (0.49) & \textbf{-0.86 (2.23)} &   -0.53 (2.03) \\
   & mat-2 & \textbf{0.79 (0.26)} &   0.88 (0.24) &   1.07 (0.16) &   1.12 (0.20) &   0.86 (0.29) &    0.83 (0.31) \\
   & mat-6 &   1.09 (0.18) &   1.31 (0.04) &   1.02 (0.21) &   1.21 (0.20) & \textbf{1.01 (0.23)} &    1.02 (0.42) \\
   & mic-10 &   1.80 (0.11) &   1.90 (0.07) &   1.75 (0.09) &   1.77 (0.10) &   1.76 (0.10) & \textbf{1.72 (0.13)} \\ 
   & mic-5 & \textbf{0.53 (0.33)} &   0.91 (0.21) &   0.68 (0.25) &   0.67 (0.38) &   0.66 (0.28) &    0.60 (0.36) \\
   & nrobot-4 &  -1.10 (0.86) &  -1.15 (1.20) & \textbf{-1.39 (0.95)} &  -0.95 (1.02) &  -0.86 (0.60) &   -1.07 (0.92) \\
\bottomrule
\end{tabular}
\caption{Mean and standard deviation of the log(regret) after 100 steps of asynchronous BO.}
\label{tab:async_results_100}
\end{table*} 
